\documentclass[letterpaper, 10 pt, conference]{ieeeconf} 
\IEEEoverridecommandlockouts             
\pdfoutput=1  

\usepackage[pdftex]{graphicx}
\usepackage{url}
\usepackage{color}
\usepackage{amsmath}

\usepackage{listings}
\usepackage[table]{xcolor}
\usepackage{tikz}
\usetikzlibrary{shapes,positioning,decorations.pathmorphing}
\usepackage{varwidth}
\usepackage{tcolorbox}
\usepackage{smartdiagram}
\usepackage{pdfpages}
\usepackage{float}

\newcommand{\craftsman}{CRAFTSMAN~} 

\definecolor{lightGray}{rgb}{0.8,0.8,0.8}

\tikzstyle{pink_block} = [rectangle, rounded corners, text width=1.5cm, minimum height=0.5cm,text centered, draw=black, fill=red!10]
\tikzstyle{white_block} = [rectangle, rounded corners, text width=3.0cm, minimum height=0.5cm,text centered, draw=black]
\tikzstyle{lilac_block} = [rectangle, rounded corners, text width=1.5cm, minimum height=0.5cm,text centered, draw=black, fill=red!30!blue!10]

\title{\LARGE \bf
ADAMANT: A Pipeline for Adaptable Manipulation Tasks
}

\author{Ana Huam\'{a}n Quispe$^{*}$ \and Stephen Hart \and Seth Gee \and Robert R. Burridge
\thanks{All authors are with TRACLabs, Inc. 16969 N. Texas Ave. Suite 300, Webster, TX. 77598. $^{*}$Corresponding author: {\tt\small ana@traclabs.com}.}%
\thanks{This work was supported by NASA contracts 80NSSC18P2203 and 80NSSC19C0216.}}

\begin{document}

\maketitle
\thispagestyle{empty}
\pagestyle{empty}

\begin{abstract}
  This paper presents ADAMANT, a set of software modules that provides grasp planning
  capabilities to an existing robot planning and control software framework. Our presented
  work allows a user to adapt a manipulation task to be used under widely
  different scenarios with minimal user input, thus reducing the operator's cognitive load.
  The developed tools include (1) plugin-based components that make it easy to extend default capabilities and to use third-party grasp libraries, (2) An object-centric way to define task
  constraints, (3) A user-friendly Rviz interface to use the grasp planner utilities, and
  (4) Interactive tools to use perception data to program a task. We tested our framework on
  a wide variety of robot simulations.
\end{abstract}

{\let\thefootnote\relax\footnote{{Preprint. Under review}}}

\section{INTRODUCTION}
Robot manipulation  applications continue to push the boundaries towards more unstructured and
unpredictable environments. Robots nowadays can be found in a wide range of
scenarios, from warehouse floors, such as Stretch \cite{ackerman2022robot},
to hospital corridors, such as Moxi \cite{ackerman2020diligent}; and for diverse
applications, from humanoids in search and rescue
situations \cite{jorgensen2019deploying}, to helpful caretakers for space missions
\cite{diftler2011robonaut}.

\begin{figure}[t!h]
\centering
\begin{tikzpicture}[framed,background rectangle/.style={thick,draw=black, rounded corners=1em, inner sep=0.05cm}]
\node[xshift=0cm,yshift=0cm]{\includegraphics[width=0.4\textwidth]{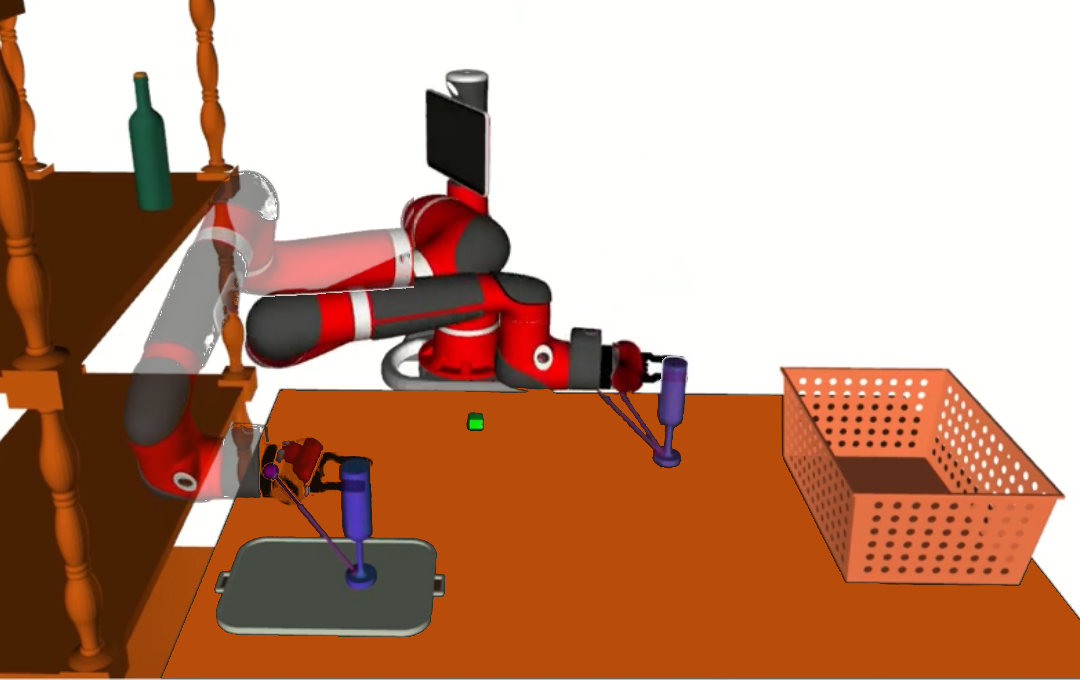} };
\node[xshift=0cm,yshift=-4.5cm]{\includegraphics[width=0.4\textwidth]{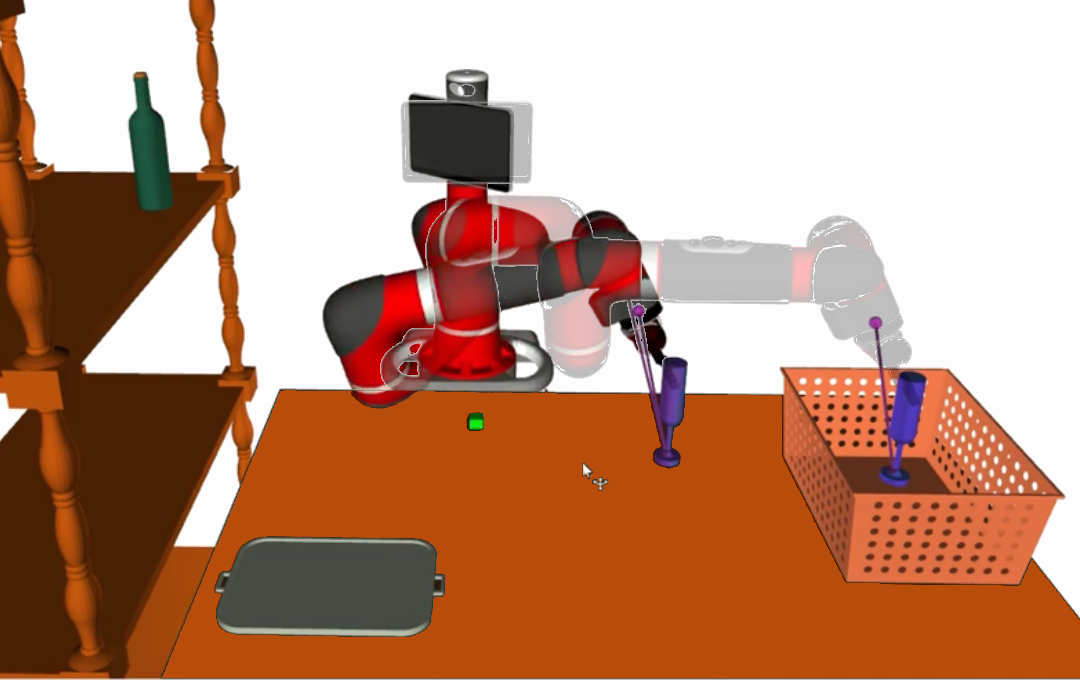} };
\node[xshift=0cm,yshift=-9cm]{\includegraphics[width=0.4\textwidth]{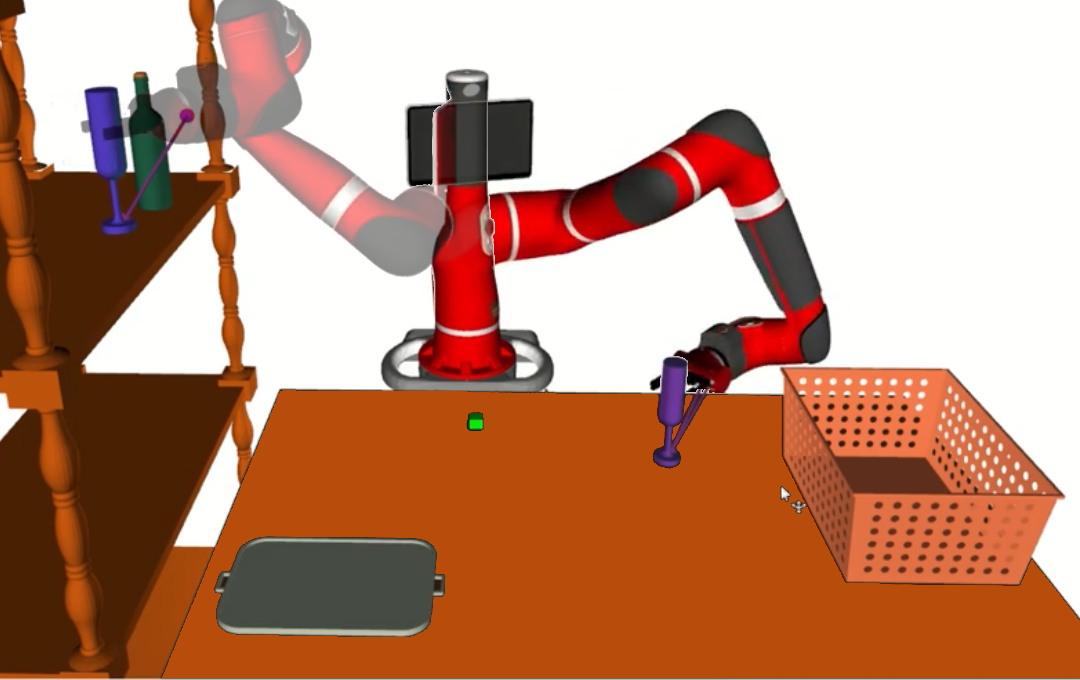} };
\node[fill=white, opacity=0.75,xshift=0cm,yshift=2cm]{\color{blue} \scriptsize (a) Grasp from the left};
\node[fill=white, opacity=0.75,xshift=0cm,yshift=-2.5cm]{\color{blue} \scriptsize (b) Grasp from above};
\node[fill=white, opacity=0.75,xshift=0cm,yshift=-7cm]{\color{blue} \scriptsize (c) Grasp from the right};
\end{tikzpicture}
\vspace*{-5pt}
\caption{Sawyer robot using 3 different grasps for 3 pick and place tasks with varying environmental constraints.}
\label{fig:cover}
\end{figure}
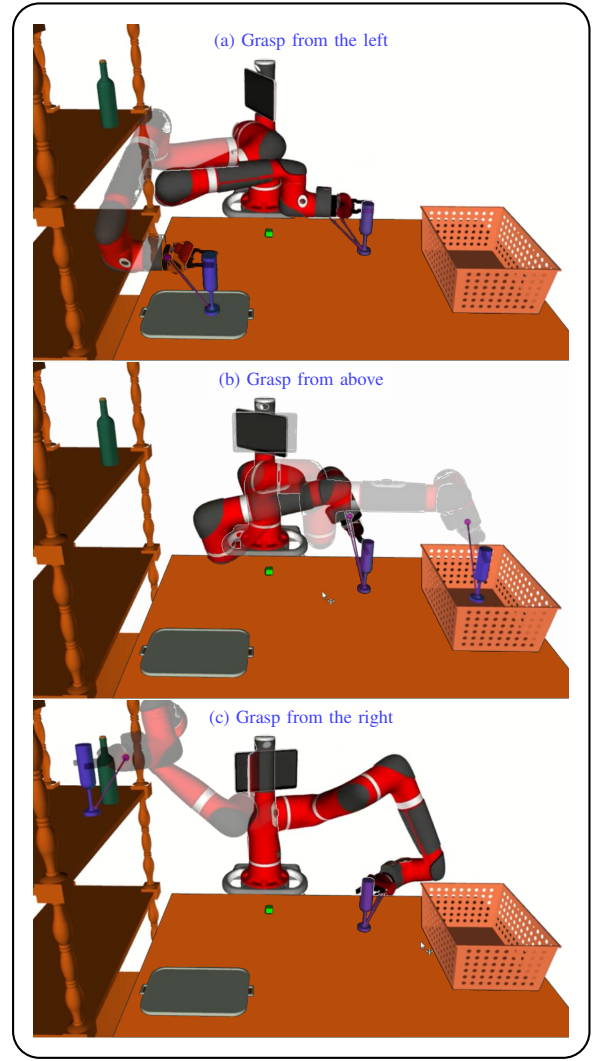

Depending on the application, input from a human operator might be needed, requiring cognitive
load of different degrees.
For applications where the robotic system is expected to operate mostly autonomously, the user's
role is limited to being supervisory in nature. On the other hand, tasks that might involve
unpredictable variables---such as changing environments or manipulating novel objects---might
require increased user input. Imagine, for instance, an example such as the one shown in Figure
\ref{fig:cover}. A Sawyer robot performs a pick-and-place task to transport a glass.
As the goal pose of the object changes, the robot must choose different grasps; e.g., for placing the glass inside the box the grasp has to be from above, whereas to place the glass in
the cabinet, the grasp has to come from the right. Having multiple grasp options thus provides
a robot with flexibility to adapt to changing scenarios. 

The work presented in this paper focuses on the development of a grasp planning software framework called ADAMANT (Adaptable Manipulation for Tasks) that allows a user to select a grasp from a set of candidates calculated on the fly by a grasp synthesis module. These candidates are visually presented via a user-friendly interface, which allows the user to quickly go through all the kinematically-feasible alternatives. Furthermore, the candidates are ordered such that the first grasps shown are the ones that maximize a user-defined metric, allowing the user to see the most promising grasps first. All our tools are designed as plugins, which permits extensibility to use third-party libraries that address  grasp generation and grasp metric implementations. Finally, our framework uses an object-centric task description language that greatly facilitates robot- (and object-) agnostic development.

Our main goal with this work was to reduce the cognitive load for a human operating a robot
performing manipulation tasks under variable environmental conditions. We present several
results in kinematic simulations that showcase the software tools developed being utilized with a variety of robots, objects, and tasks with varying degrees of complexity. 

The rest of this paper is organized as follows. Section \ref{sec:related_work} describes
related work on currently available grasp planning frameworks, as well as tools for
grasp synthesis and evaluation. Section \ref{sec:task_description} succinctly defines
the object-centric task description we use as input to our grasp planning modules. Section \ref{sec:arch} presents the architecture of our developed software and its integration to our existing
software framework for robot planning and control. Section \ref{sec:interfaces} presents further
details on the UI interface for grasp generation using Rviz and a widget for easily capturing
RGB-D sensor data and incorporate it into the grasp planning pipeline. Section \ref{sec:examples} shows some of the demonstrations performed in a variety of robot kinematic simulations. Finally, Section \ref{sec:future_work} summarizes our conclusions as well as proposed future work.

\section{Related Work}
\label{sec:related_work}

\subsection{Grasp Planning}
Work on grasp planning is vast and covers a variety of approaches that have dramatically
evolved as robot hardware and computational resources have become more accessible. Classical
approaches primarily consisted of the generation of robot hand poses and the use of analytic
metrics, such as the force-closure metric $\epsilon$ proposed by Ferrari and Canny \cite{ferrari1992planning} to determine whether the grasp was valid or not. Hand and finger poses have been generated by a range of strategies, including optimization approaches \cite{ciocarlie2007dimensionality}, surface sampling \cite{diankov2008openrave}, primitive shape decomposition \cite{huebner2008selection,przybylski2010unions,goldfeder2007grasp} and heuristic approaches \cite{balasubramanian2014physical,ciocarlie2014towards}. During the last few years, the advances of deep learning, and particularly, its application on data such as point clouds, have produced a number of approaches, such as 6DOF-Graspnet \cite{mousavian20196} and derived approaches (e.g., \cite{murali20206}) that use the Pointnet++ architecture \cite{qi2017pointnet++}.

Most of the aforementioned approaches focus on generating a grasp for, primarily, \textit{just grasping} the object. That is, picking it up from its original location and guaranteeing that it can withstand some mild perturbation. While this is useful for simple transport tasks, it is not sufficient for tasks that consist of more than one step (e.g., rotating a wheel, opening a door, pouring a drink, painting on a planar surface). MoveIt! \cite{sucan2013moveit}, a very popular robot planning framework, provides Rviz tools that allow a user to define tasks by means of their Task Constructor module, and
generate grasps using third-party libraries such as GPD and DexNet. While our presented framework offers similar capabilities to those of MoveIt!, it also presents some features that make it more extensible (Section \ref{subsec:grasp_planner}), and allows a more interactive way for a user to be involved in the grasp planning process (Section \ref{sec:interfaces}). Furthermore,  the task description language we use is fairly simple (Section \ref{subsec:at_tdl}) and is easily editable by the user either manually (directly modifying a single JSON file) or interactively, within the Rviz 3D window.

\subsection{CRAFTSMAN}
\label{subsec:background_craftsman}
\begin{figure}[h!]
  \centering
  \vspace*{-10pt}
  \includegraphics[width=.925\linewidth]{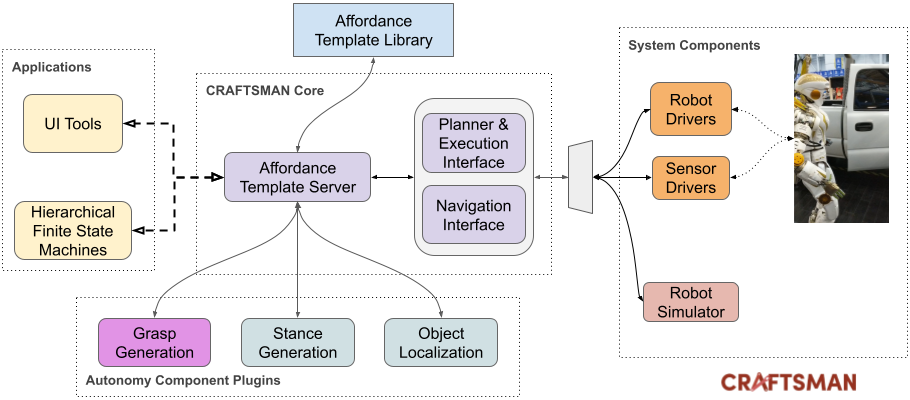}
  \vspace*{-5pt}
  \caption{A high-level overview of the commercially-deployed \craftsman framework, which has been demonstrated on a variety of industrial manipulators, research robots, and humanoid platforms.}
  \label{fig:craftsman_overview}
\end{figure}

Over the past 7 years, TRACLabs researchers have developed a powerful suite of robot planning, control, and tasking software called \craftsman (Figure~\ref{fig:craftsman_overview}).  \craftsman is designed to be robot agnostic, providing generic collision-free path planning, trajectory optimization, action-sequencing capabilities, and methods for parameterizing, encoding, and visualizing task descriptions.  One innovation of \craftsman is that it enables users to create robot-independent task descriptions that can be re-used across deployments and can endow robot systems with a set of core competencies.  This has allowed TRACLabs to successfully demonstrate \craftsman components on a variety of robotic platforms~\cite{james2015prophetic, jorgensen2019deploying}.

\craftsman provides a straightforward tool suite for both expert and non-expert developers~\cite{beeson2016cartesian}.  The current software implementation uses libraries from the {\it Robot Operating System~(ROS)} ecosystem, including inter-process messaging and Rviz 3D visualization tools ~\cite{gossow11interactive}, providing easy integration with custom or off-the-shelf perception and navigation packages for systems that have 3D sensing and/or mobile capabilities.  In particular, \craftsman contains an implementation of a graphical task description language called the {\it Affordance Template} framework~\cite{hart2015affordance, hart2022affordance} that greatly simplifies task design.

\begin{figure*}[t!]
\centering
\begin{tikzpicture}[framed,background rectangle/.style={thick,draw=black, rounded corners=1em, inner sep=0.05cm}]
\node[xshift=0cm,yshift=0cm]{\includegraphics[height=4.0cm]{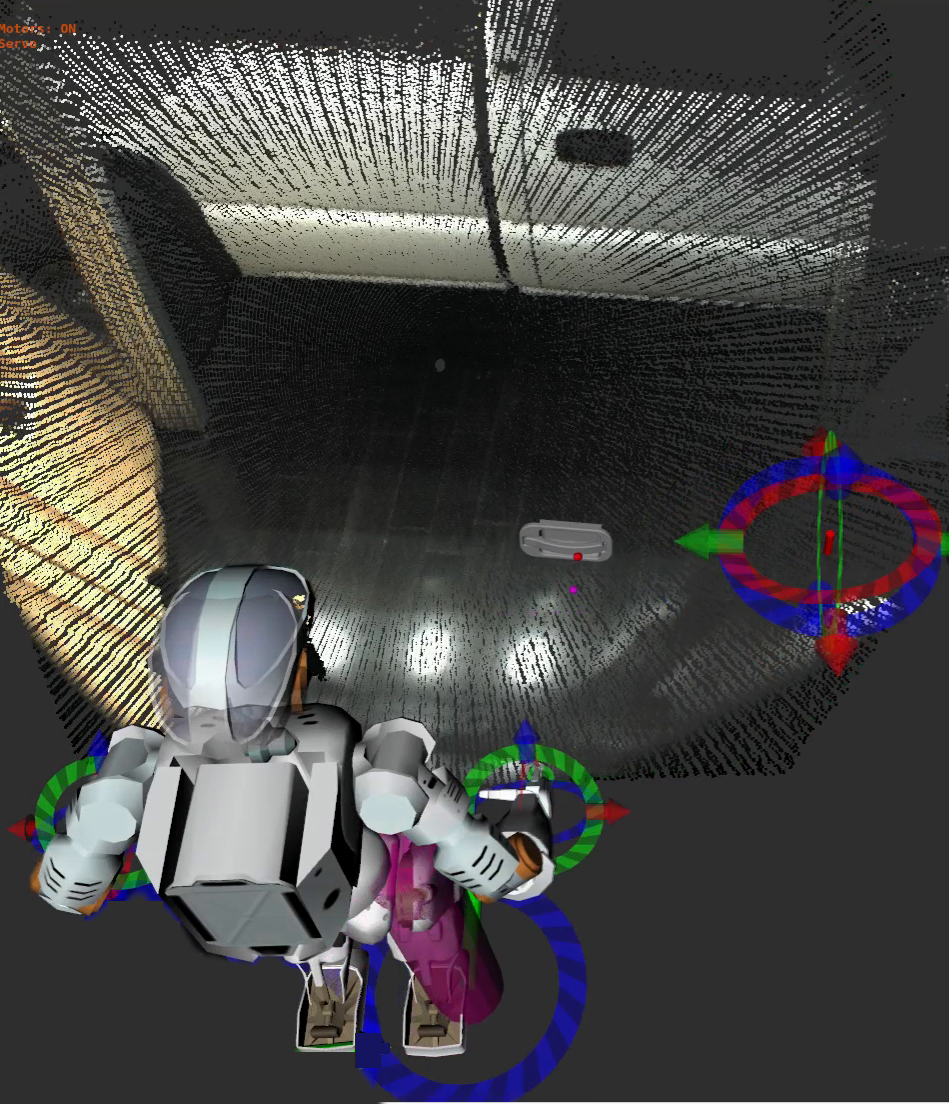} };
\node[xshift=3.5cm,yshift=0cm]{\includegraphics[height=4.0cm]{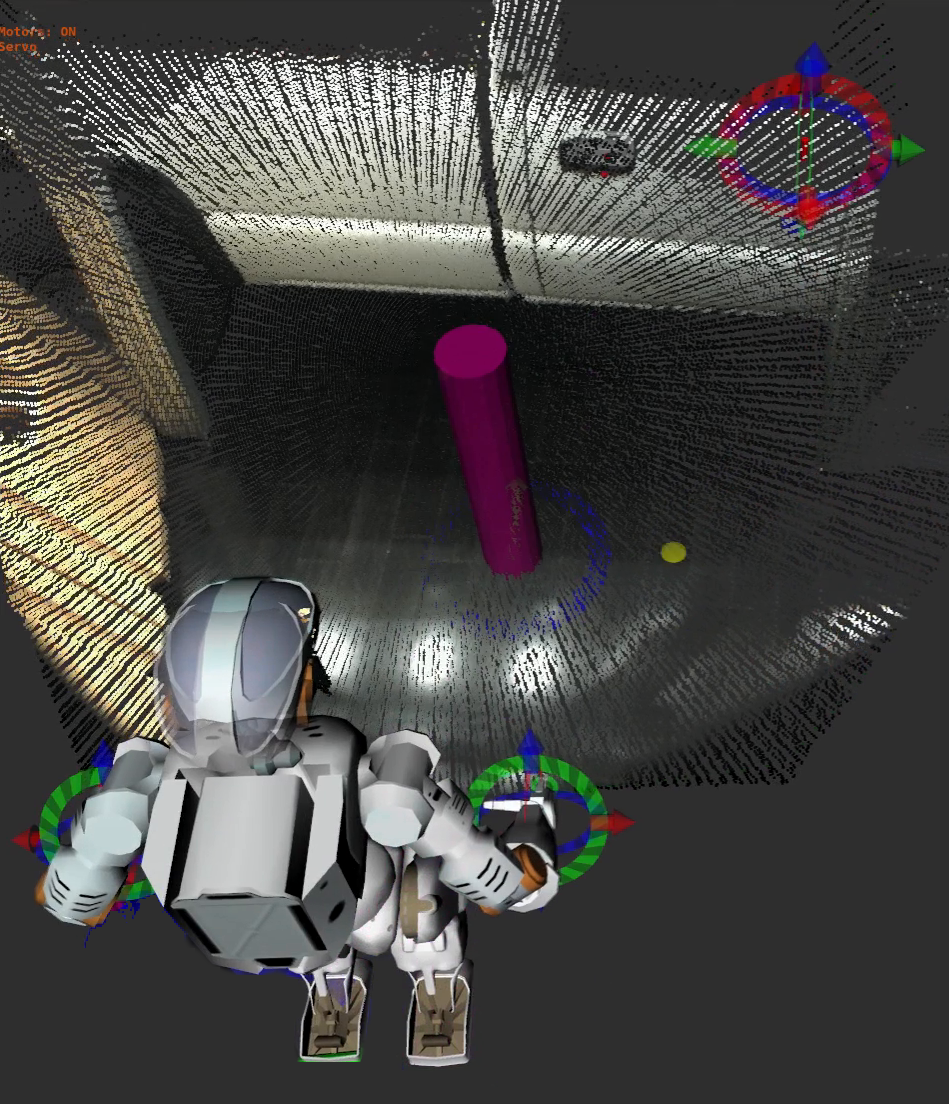} };
\node[xshift=6.9cm,yshift=0cm]{\includegraphics[height=4.0cm]{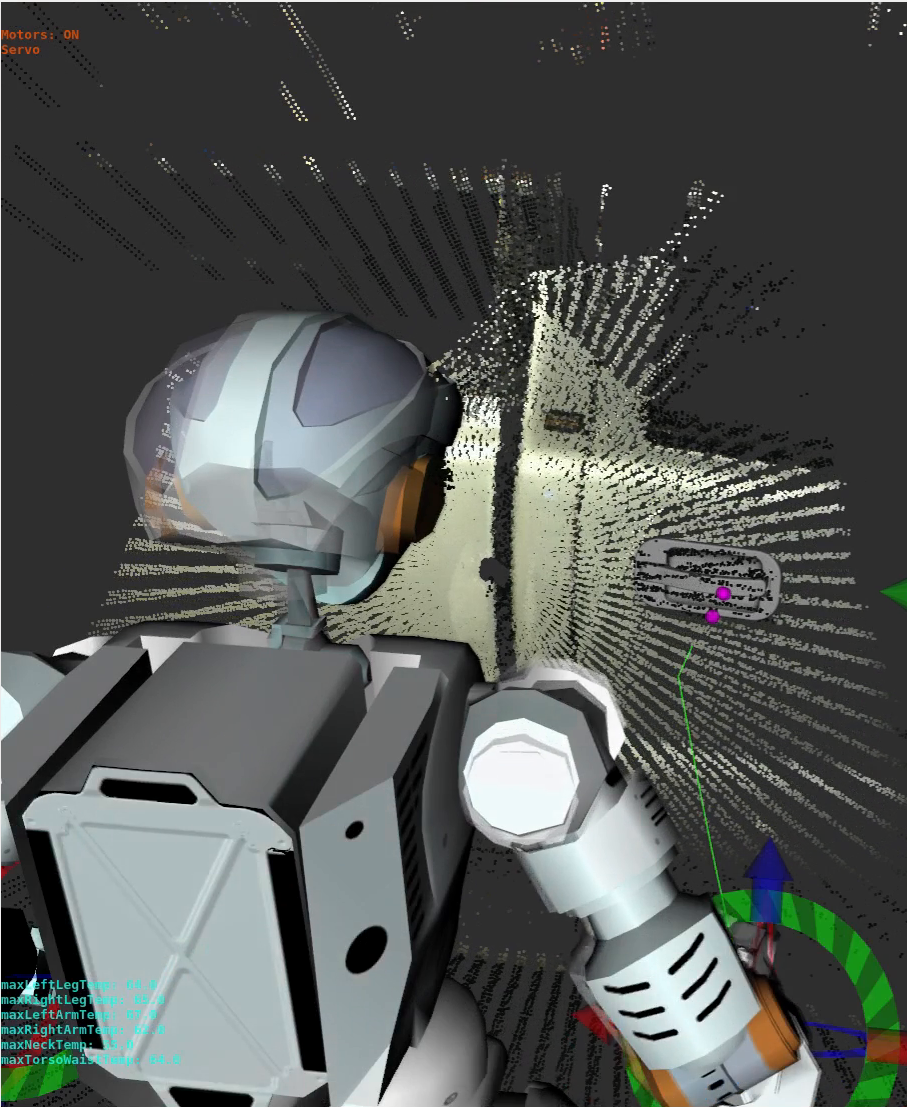} };

\node[xshift=7.85cm,yshift=1.45cm]{\includegraphics[height=1.0cm]{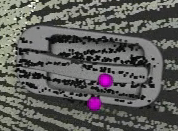} };
\draw [magenta,opacity=0.75, ultra thick] (7.2cm,1.0cm) rectangle (8.5cm, 1.95cm);
\draw[magenta, very thick, -stealth] (7.8cm,0.2cm) -- (7.8cm, 1.0cm);

\node[xshift=10.325cm,yshift=0cm]{\includegraphics[height=4.0cm]{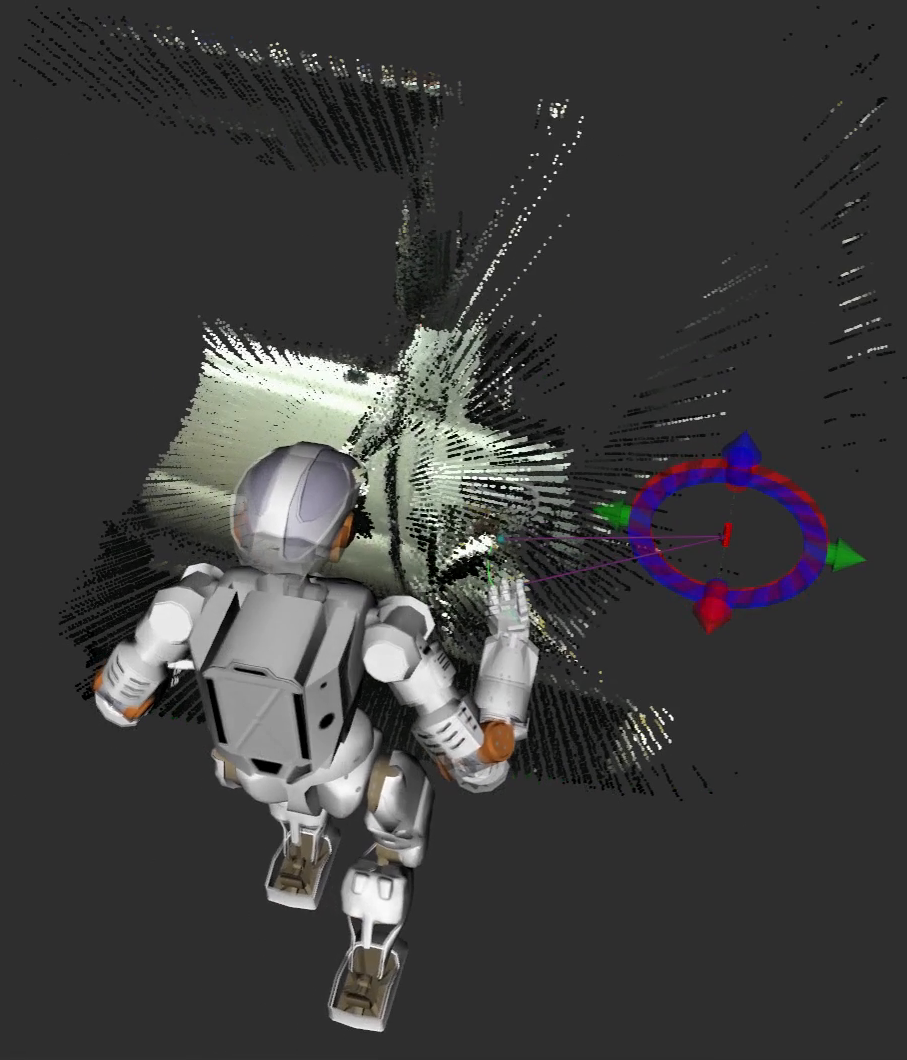} };
\node[xshift=13.25cm,yshift=0cm]{\includegraphics[height=4.0cm]{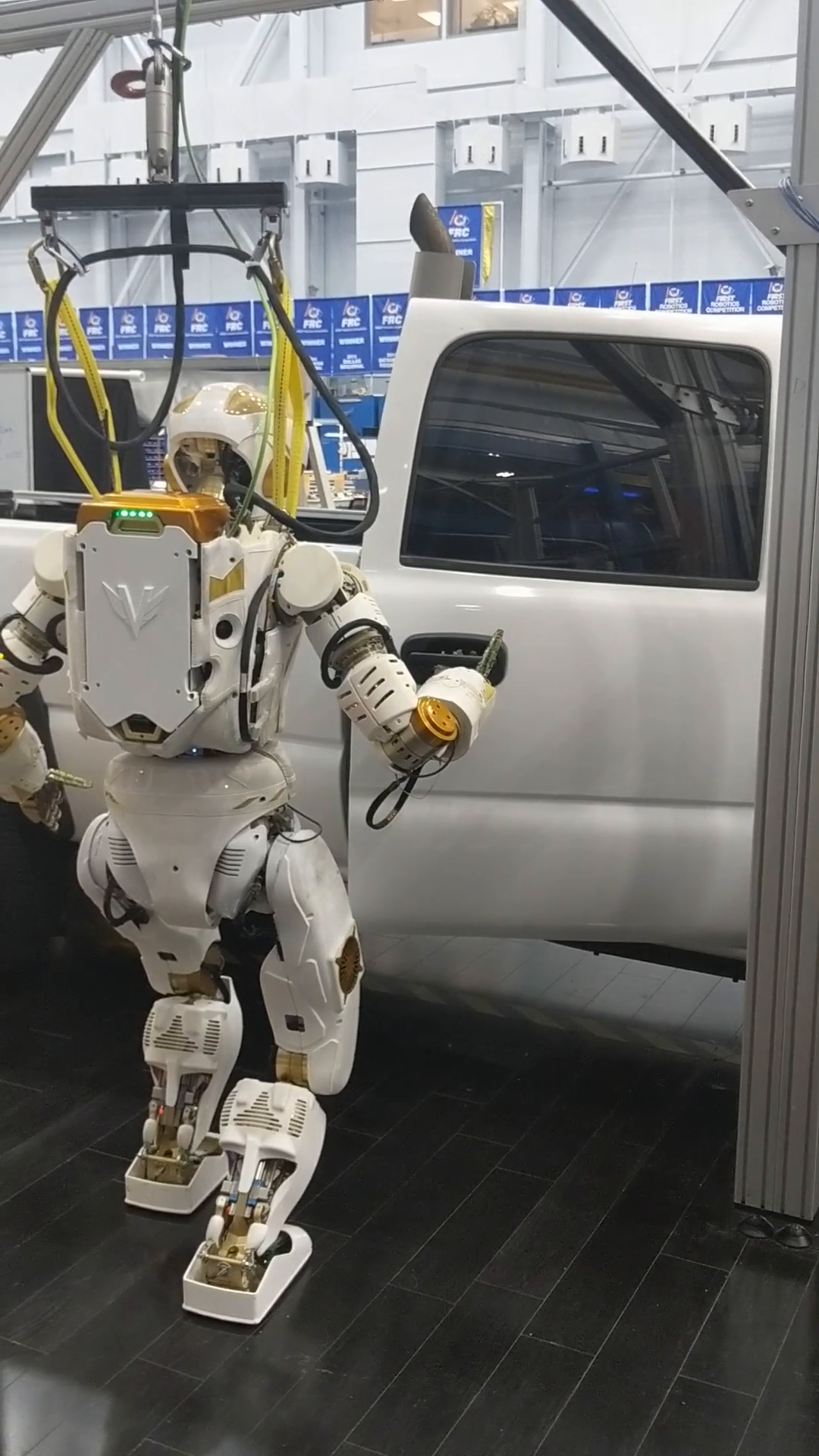} };
\node[fill=white,xshift=-1.41cm,yshift=-1.8cm]{\color{blue} (a)};
\node[fill=white,xshift=2.1cm,yshift=-1.8cm]{\color{blue} (b)};
\node[fill=white, opacity=1.0, xshift=5.55cm,yshift=-1.8cm]{\color{blue} (c)};
\node[fill=white, xshift=8.9cm,yshift=-1.8cm]{\color{blue} (d)};
\node[fill=white, xshift=12.4cm,yshift=-1.8cm]{\color{blue} (e)};
\node[xshift=4.4cm,yshift=0.3cm]{\color{magenta} Nav wp};
\draw[draw, fill=green!50, opacity=0.25] (0.35cm, -0.02cm) circle (0.4cm);
\draw[draw, fill=green!50, opacity=0.25] (4.0cm, 1.5cm) circle (0.4cm);
\end{tikzpicture}
\vspace*{-5pt}
\caption{Affordance Template usage example: Performing a mobile manipulation task to approach and open a door. (a) Start configuration. (b) User aligned AT with sensor data (moved door handle IM marker -highlighted with green- to match cloud). (c) Robot navigated towards the Navigation waypoint IM, EE waypoints (closeup in top-right corner) are now reachable, (d) Robot opened the  door moving its hand through the EE waypoints, (e) View of real robot after (d).}
\label{fig:at_example}
\end{figure*}
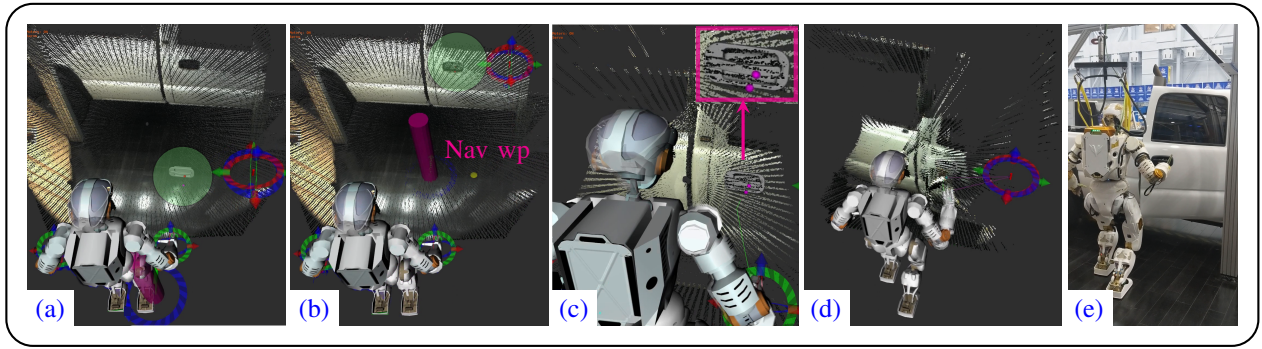

\subsection{The AT Task Description Language}
\label{subsec:at_tdl}
The Affordance Template~(AT) specification is a {\em task description language} that provides definitions for tasks that can be used by a variety of robots in a variety of contexts.  Specifically, ATs allow a programmer to specify sequences of navigation and end-effector waypoints represented in the coordinate systems of stationary or manipulable objects in the environment, including the robot itself when necessary.  Virtual overlays of desired objects and task-specific end-effector waypoints can be visualized and adjusted in a 3D GUI, which also contains sensory data and related environmental models.  Waypoints can include conditioning information, such as the type of path planning to use or the style of manipulation motion to prefer (e.g., straight-line Cartesian versus joint motion).  Waypoints can also specify attached objects, so that grasped objects will be accounted for automatically when checking for collisions during planning.

Figure~\ref{fig:at_example} shows an example where an operator registers an open-door affordance template to a robot's 3D sensor data.  The lower levels of the \craftsman framework then use the updated AT waypoints to plan arm and (if required) navigation trajectories.

Although CRAFTSMAN has been used successfully in a variety of robot applications through the years, it lacks autonomous
grasp planning capabilities. For instance, for the example shown, the hand pose that allows the robot to
grasp the door was crafted by the user, with the help of the Affordance Template Rviz interactive tools. The goal of the ADAMANT modules was to provide CRAFTSMAN with those grasping capabilities. The software modules we developed and present in the rest of this paper allow CRAFTSMAN users to use a grasp planning module to generate
grasps automatically, thus making the task of designing manipulation tasks easier for a developer.
\section{Task Representation}
\label{sec:task_description}
In this section, we describe the low-level task representation we utilize to represent a robot manipulation problem \textit{within the grasp planning module}, specifically, its input (task description) and output (candidate grasps):



\subsection{Input: Task Description}
\begin{figure}[t!h]
\centering
\begin{tikzpicture}[framed,background rectangle/.style={thick,draw=black, rounded corners=1em, inner sep=0.05cm}]
\node[xshift=0cm,yshift=0cm]{\includegraphics[width=0.23\textwidth]{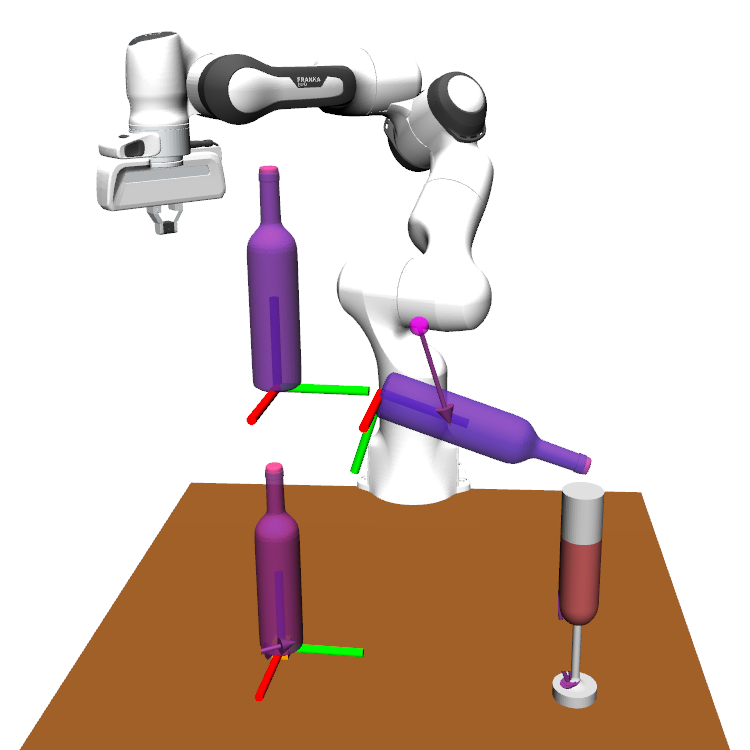} };
\node[xshift=4cm,yshift=0cm]{\includegraphics[width=0.23\textwidth]{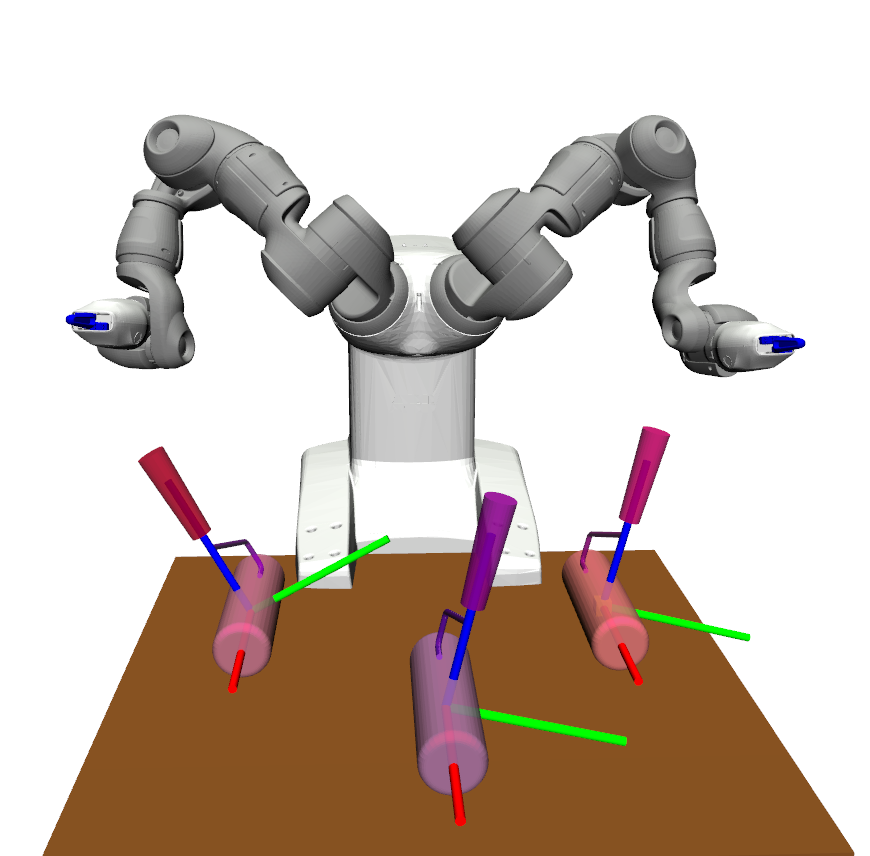} };
\node[fill=white, opacity=0.75,xshift=0cm,yshift=2cm]{\color{blue} \scriptsize (a) Pour task steps};
\node[fill=white, opacity=0.75,xshift=4cm,yshift=2cm]{\color{blue} \scriptsize (b) Paint task steps};
\end{tikzpicture}
\vspace*{-5pt}
\caption{Visualization of object-centric task step definition (Object Waypoints). These are Rviz IMs that allow a user to edit them interactively (right-click to make 6D gimbal visible).}
\label{fig:task_description}
\end{figure}

We represent a manipulation problem with a custom ROS message named \texttt{FullTaskDescription}, shown in Fig. \ref{fig:full_task_description}. It contains:

\begin{itemize}
\setlength{\itemsep}{0pt}%
\item{\textit{MinimalTaskDescription:} The geometric information of the object to be manipulated and the name of the end effector to be used.}  
\item{\textit{Task steps:} Discrete 3D poses (with corresponding tolerances) that represent the object pose at relevant points. Examples of those are shown in Fig. \ref{fig:task_description}. The tolerances are expressed with respect to the object's frames of reference.}  
\end{itemize}

\begin{figure}[h]
\centering
\begin{tcolorbox} [width=8cm]
\begin{lstlisting}
# Minimal task description
string ee_group
CollisionObjectInfo object
  
#Steps
geometry_msgs/PoseStamped[] steps
geometry_msgs/Vector3[] tol_pos
geometry_msgs/Vector3[] tol_rot

# Arm start pose
sensor_msgs/JointState start_arm_config   
\end{lstlisting}
\end{tcolorbox}
\vspace{-2ex}
\caption{FullTaskDescription.msg}
\label{fig:full_task_description}
\end{figure}




Note that our task description (for grasp planning purposes) does not include information about \textbf{path constraints}, if any;
this is due to the fact that while grasp planning is expected to return grasps that are kinematically reachable at each task step, it is  \textbf{not} expected to perform motion planning for moving the arm between steps, as this might be too time-consuming, especially if dozens of grasps are generated and a task contains multiple steps. Finally, note that this Task Description representation is used internally by the grasp planning module. The user only has to deal with the Affordance Template default representation. Any required transformation between message types (e.g. AT Object Waypoints into Task steps) is done in the background.

\subsection{Output: Candidate grasp set}
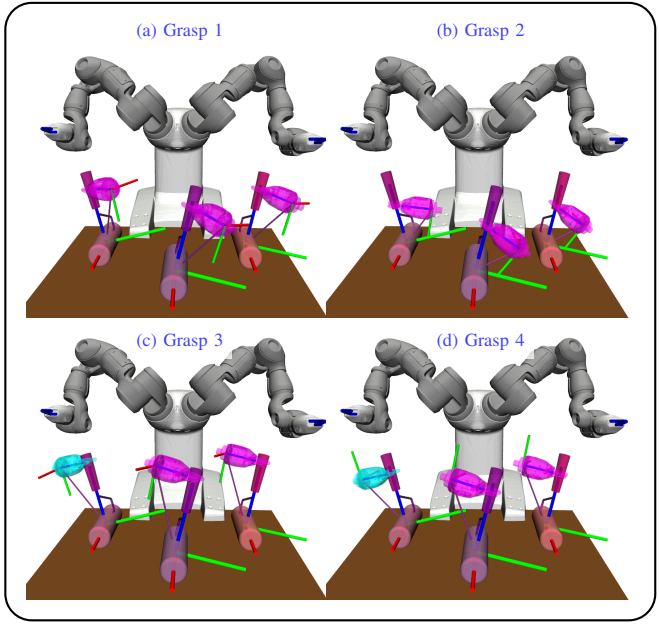
\begin{figure}[t!h]
\centering
\begin{tikzpicture}[framed,background rectangle/.style={thick,draw=black, rounded corners=1em, inner sep=0.05cm}]
\node[xshift=0cm,yshift=0cm]{\includegraphics[width=0.23\textwidth]{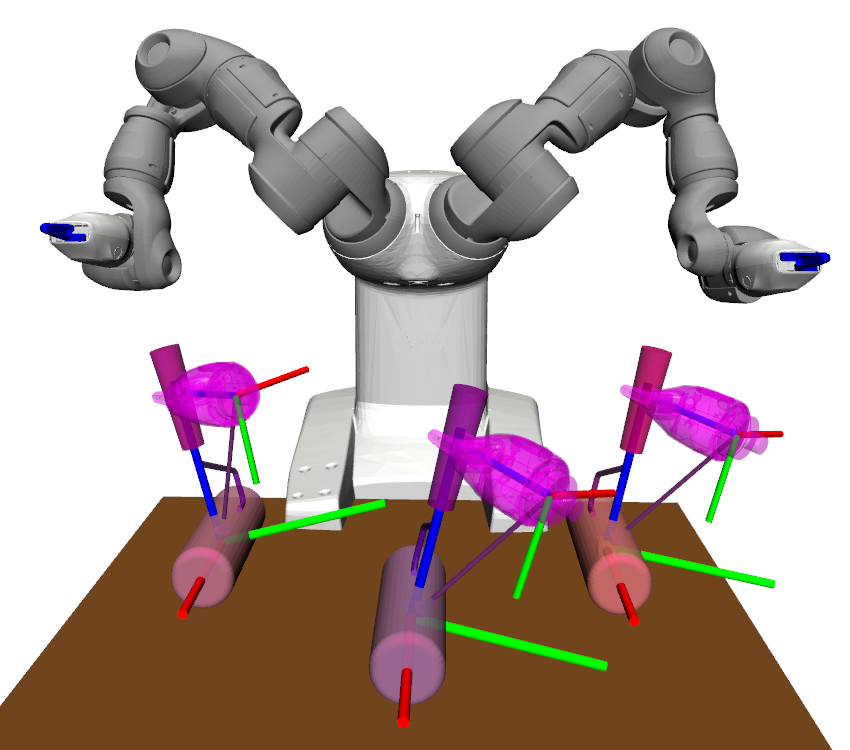} };
\node[xshift=4cm,yshift=0cm]{\includegraphics[width=0.23\textwidth]{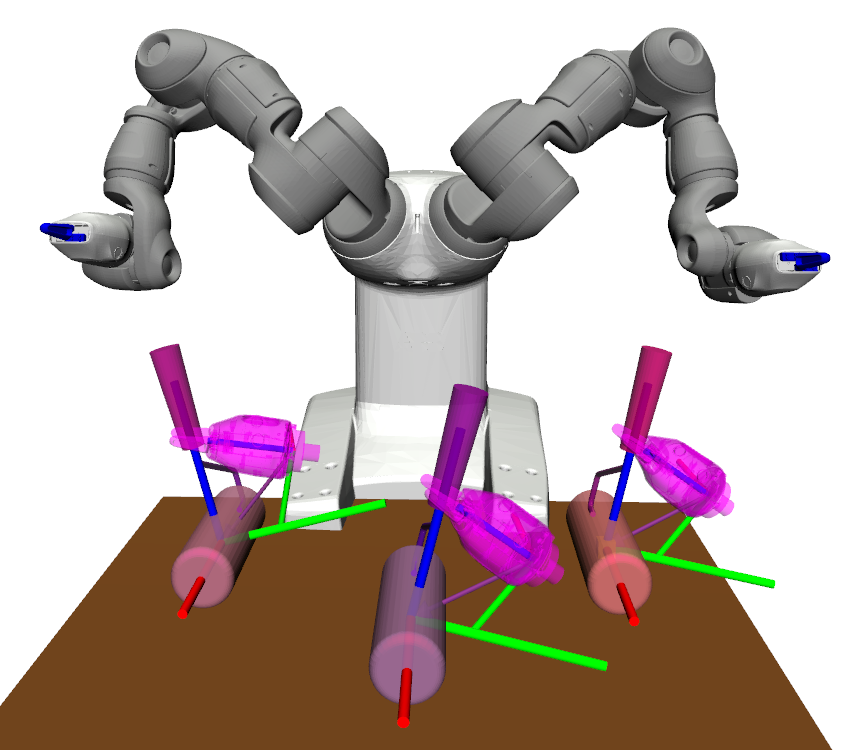} };
\node[xshift=0cm,yshift=-3.75cm]{\includegraphics[width=0.23\textwidth]{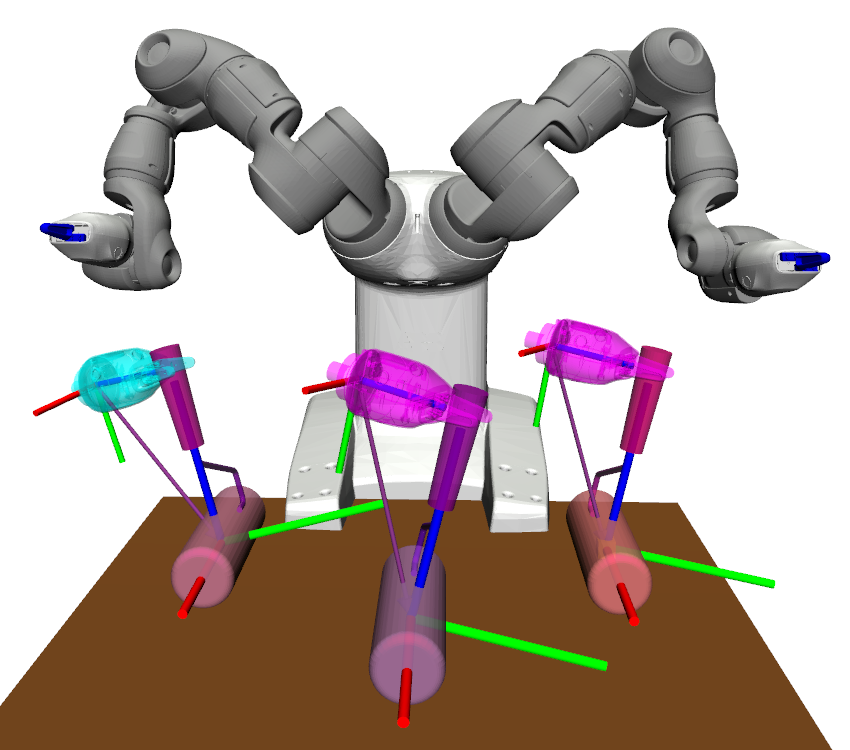} };
\node[xshift=4cm,yshift=-3.75cm]{\includegraphics[width=0.23\textwidth]{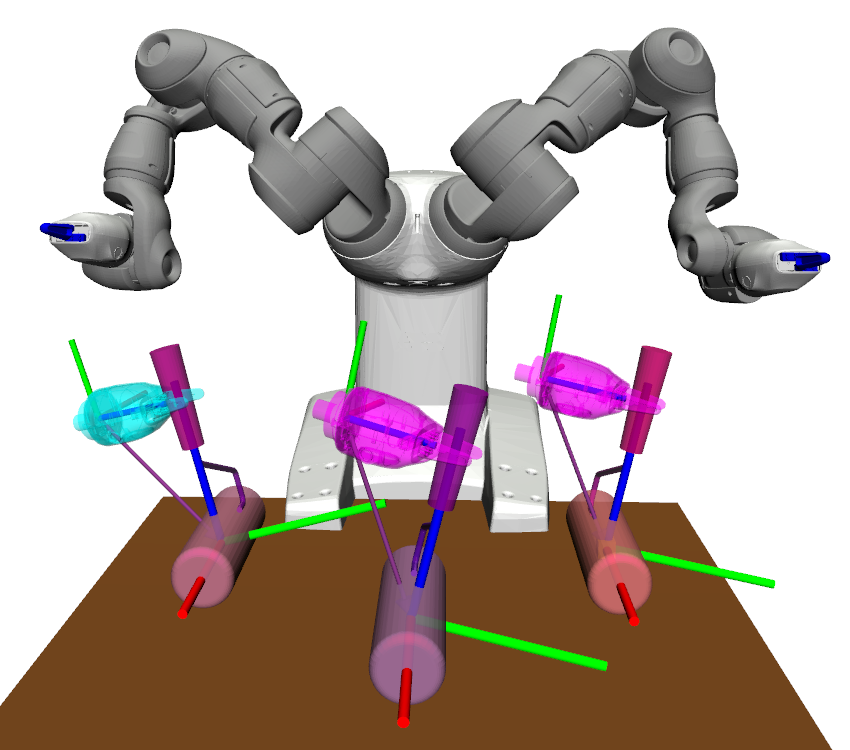} };

\node[fill=white, opacity=0.75,xshift=0cm,yshift=2cm]{\color{blue} \scriptsize (a) Grasp 1};
\node[fill=white, opacity=0.75,xshift=4cm,yshift=2cm]{\color{blue} \scriptsize (b) Grasp 2};
\node[fill=white, opacity=0.75,xshift=0cm,yshift=-2.125cm]{\color{blue} \scriptsize (c) Grasp 3};
\node[fill=white, opacity=0.75,xshift=4cm,yshift=-2.125cm]{\color{blue} \scriptsize (d) Grasp 4};
\end{tikzpicture}
\vspace*{-5pt}
\caption{Visualization of output (some of the grasp candidates used to update AT End Effector Waypoints). The magenta markers represent the pose of the end effector with respect to the target object. The cyan color indicates that the grasp at this step is only reachable by considering the tolerances}
\label{fig:grasps}
\end{figure}

The output, or solution to a robot manipulation problem is a set of candidate grasps. A grasp is defined as (1) the pose of the hand's TCP (tool center point) with respect to the object's frame, (2) the finger joint configuration that positions the fingers in contact with the object, and (3) the name of the end effector. Figure \ref{fig:grasps} shows some of the candidate grasps generated for a painting task.

\section{Architecture}
\label{sec:arch}
\begin{figure}[h!]
\centering
\begin{tikzpicture}[framed,background rectangle/.style={thick,draw=black, rounded corners=1em, inner sep=0.05cm}]
\node[xshift=0cm,yshift=0cm]{\includegraphics[width=0.45\textwidth]{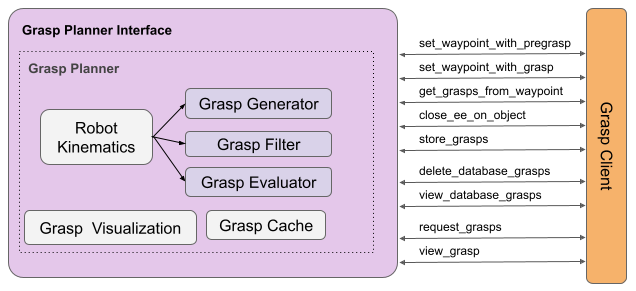} };
\end{tikzpicture}
\vspace*{-7pt}
\caption{Diagram of the ADAMANT Grasp Planning components.}
\label{fig:adamant_arch}
\end{figure}

The ADAMANT structure is shown in Figure \ref{fig:adamant_arch}. The most important
component of these modules is the \textbf{Grasp Planner}, which we explain in some more detail as follows.

\subsection{Grasp Planner}
\label{subsec:grasp_planner}
Under our implementation, the grasp planning process consists of three sequential steps, each of which is operated by the modules described below: 

\begin{enumerate}
\setlength{\itemsep}{0pt}%
\item{\textbf{Grasp Generator:} This module receives the task description and uses the object's geometry and robot hand's information to generate a set of candidate grasps that allow the robot hand to close around the object and have at least two points of contact with it.} 
\item{\textbf{Grasp Filter:} This module receives the candidate grasps produced by the Grasp Generator and only keeps the candidates that are kinematically feasible to execute at all the manipulation task steps with tolerances.}
\item{\textbf{Grasp Evaluator:} This module is fed with the surviving grasps from the Grasp Filter and applies a function which calculates a heuristic measure of how \textit{adequate} a grasp is for a specific task. This metric could consider factors such as kinematic reachability (how easy it is for an arm to reach the end-effector poses at relevant steps in the task) and grasp suitability (how robust a grasp is).}  
\end{enumerate}

The output of this 3-step process is an ordered list of candidate grasps that can now be returned
to the operator, who can visualize them and select one of them, or alternatively, can rely on the
grasp metric ordering and automatically choose the first grasp in the list.


The 3 modules described are implemented using the ROS C++ \texttt{pluginlib} library, which allows the user ample flexibility to create instances of one or all of the 3 grasp planner components and seamlessly plug them in the grasp planner. This approach allows us to experiment with different grasp generation algorithms as well as diverse grasp metrics, either by encoding new grasp algorithms in these instances, or if available, taking advantage of existing third-party libraries available in the community by creating a new plugin for them. We developed a number of plugins, including the following:

\begin{itemize}
\setlength{\itemsep}{0pt}%
\item{\textbf{Default Grasp Generator:} This plugin generates candidate grasps by uniformly sampling the surface of the object and using the normals as the candidate grasp approach directions. The fingers are then moved small steps from their open to their close pose, each link stops moving if a contact is made.}
\item{\textbf{GPD Grasp Generatorn:} This plugin implements a wrapper around the popular Grasp Pose Detection library \cite{ten2017grasp} (with OpenVINO extensions) to generate grasps for prismatic grippers.}
\item{\textbf{Capability Index Grasp evaluator:} We implemented the metric proposed in \cite{mavrakis2016task}, which maximizes the projection of the task ellipsoid along the trajectory of the object. This is the metric used for most of the examples presented in this paper.}
\item{\textbf{Default Grasp evaluator:} This plugin implements the metric proposed in \cite{huaman2015thesis}, and it combines grasp and arm kinematics factors to rank grasps.}  
\end{itemize}


\subsection{Application usage}
\label{subsec:application}
We provide a number of utilities built over the core GraspPlanner class such that they facilitate
its usage. These include:
\begin{itemize}
\setlength{\itemsep}{0pt}%
\item{\textbf{Grasp Visualization:} A class that contains an Interactive Marker server and is used to display visualization of the candidate grasps. The Grasp Planner contains an instance of this class.}
\item{\textbf{Grasp Cache:} A class also contained within the Grasp Planner, which allows a user to store the grasps generated either on disk or only during the current run. The Grasp Cache uses the MinimalDescription message (which contains the combination of hand and object geometries) as a key; and the  grasps produced by the Grasp Generator step as the value. This means that during a run, the planner only calls the Grasp Generator step at most once, for a  given object and end effector combination. This is particularly useful for objects with complex geometries. The Grasp Filter and Evaluator are always called, as their filtering effect depends on the (likely changing) environment.}  
\item{\textbf{Grasp Planner Interface:} Wrapper class that contains an instance of the GraspPlanner and a set of servers providing ROS services for easy communication between the planner and external applications.}
\item{\textbf{Grasp Planner Client: } A class that contains functions that allow a user to easily request grasp planning information. In the background, these functions use the services provided by the Interface and send requests and receive responses to grasp queries.}     
\end{itemize}


\section{User Interfaces}
\label{sec:interfaces}
In addition to the tools described in  \ref{subsec:application}, which are useful
for interacting with the grasp planning capabilities in a programmatic way, we also developed
a front-end interface that allows the user to build, modify, and update manipulation tasks on the fly
using interactive Rviz tools. The following subsections summarize their usage.

\subsection{Grasping Tab}
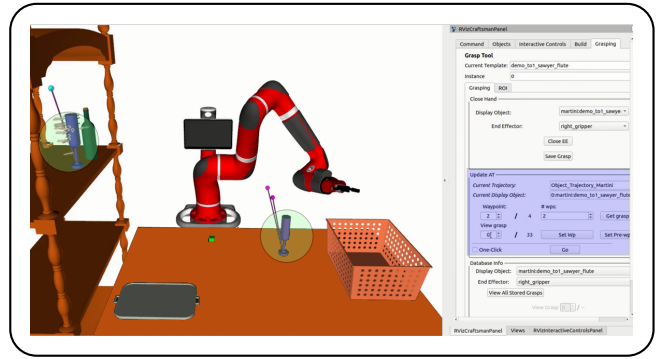
\begin{figure}[h!]
\centering
\begin{tikzpicture}[framed,background rectangle/.style={thick,draw=black, rounded corners=1em, inner sep=0.05cm}]
\node[xshift=0cm,yshift=0cm]{\includegraphics[width=0.45\textwidth]{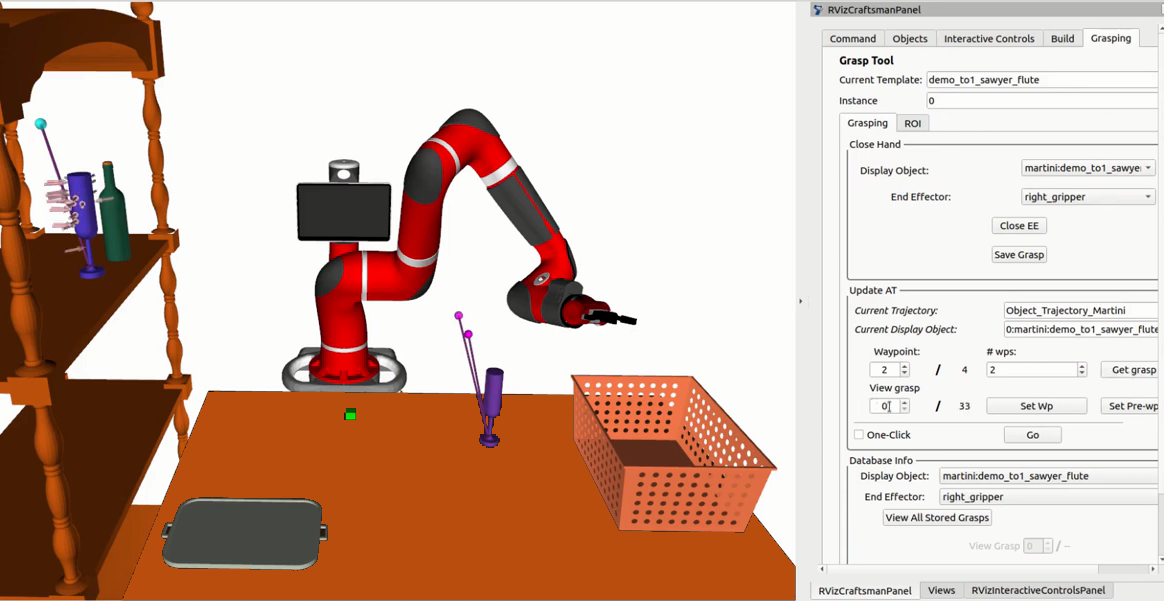} };
\draw [fill=blue,opacity=0.2] (1.8cm,0.15cm) rectangle (3.95cm,-1.0cm);
\draw [fill=green!50,opacity=0.25] (-0.6cm,-0.75cm) circle (0.35cm);
\draw [fill=green!50,opacity=0.25] (-3.5cm,0.5cm) circle (0.4cm);
\end{tikzpicture}
\vspace*{-7pt}
\caption{Grasping Tab after the \texttt{Get grasps} button has been pressed. Grasping Tab is at the right of the image. The Update AT widget (used to send grasp queries and update the AT on the fly) is highlighted in blue. The Interactive Markers for the task steps (Object waypoints) are highlighted in green.}
\label{fig:ui_grasp_tab_1}
\end{figure}

The purpose of this tab is to allow the user an easy way to calculate candidate grasps for a task defined
with Affordance Templates, our default task description language. The widget is shown in Fig. \ref{fig:ui_grasp_tab_1}\footnote{Video of a user utilizing this Tab is available here: \url{https://youtu.be/ECK4io6KWww}.}.

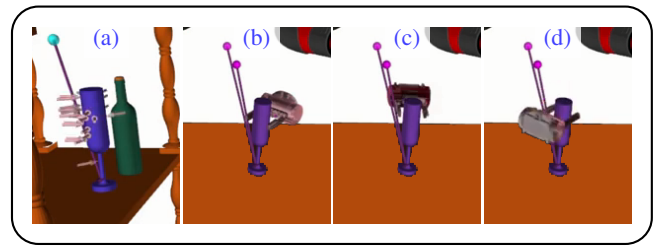
\begin{figure}[h!]
\centering
\begin{tikzpicture}[framed,background rectangle/.style={thick,draw=black, rounded corners=1em, inner sep=0.05cm}]
\node[xshift=0cm,yshift=0cm]{\includegraphics[width=0.11\textwidth]{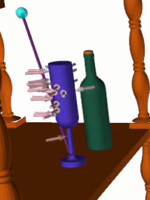} };
\node[xshift=2cm,yshift=0cm]{\includegraphics[width=0.11\textwidth]{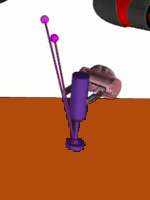} };
\node[xshift=4cm,yshift=0cm]{\includegraphics[width=0.11\textwidth]{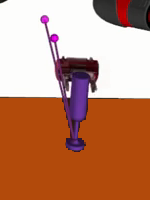} };
\node[xshift=6cm,yshift=0cm]{\includegraphics[width=0.11\textwidth]{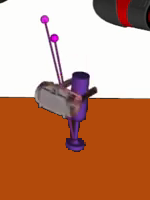} };
\node[fill=white, opacity=0.75,xshift=0.0cm,yshift=1.15cm]{\color{blue} \small (a)};
\node[fill=white, opacity=0.75,xshift=2.0cm,yshift=1.15cm]{\color{blue} \small (b)};
\node[fill=white, opacity=0.75,xshift=4.0cm,yshift=1.15cm]{\color{blue} \small (c)};
\node[fill=white, opacity=0.75,xshift=6.0cm,yshift=1.15cm]{\color{blue} \small (d)};
\end{tikzpicture}
\vspace*{-7pt}
\caption{(a) Compact (arrow) view of all the valid candidate grasps for the task. (b)-(d) Non-compact view of individual candidate grasps, available by scrolling through the spin box in the \texttt{Update AT} widget.}
\label{fig:ui_grasp_tab_2}
\end{figure}

As can be seen in the image, the user can make a grasp query with the button \texttt{Get grasp}.
If candidate grasps are returned, they can be easily viewable using the \texttt{View grasp} scroll box (Fig.\ref{fig:ui_grasp_tab_2}). The button \texttt{Set wp} is used to effectively update the current AT's End Effector waypoints with the currently selected grasp candidate. 

\subsection{Supporting Object-agnostic tasks}
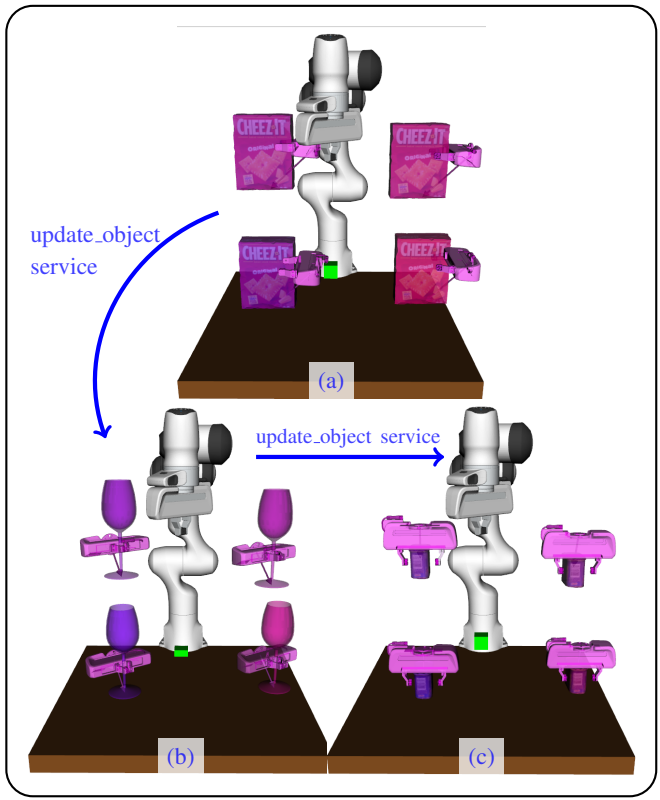
\begin{figure}[h!]
\centering
\begin{tikzpicture}[framed,background rectangle/.style={thick,draw=black, rounded corners=1em, inner sep=0.05cm}]
\node(1a)[xshift=2cm,yshift=0cm]{\includegraphics[width=0.23\textwidth]{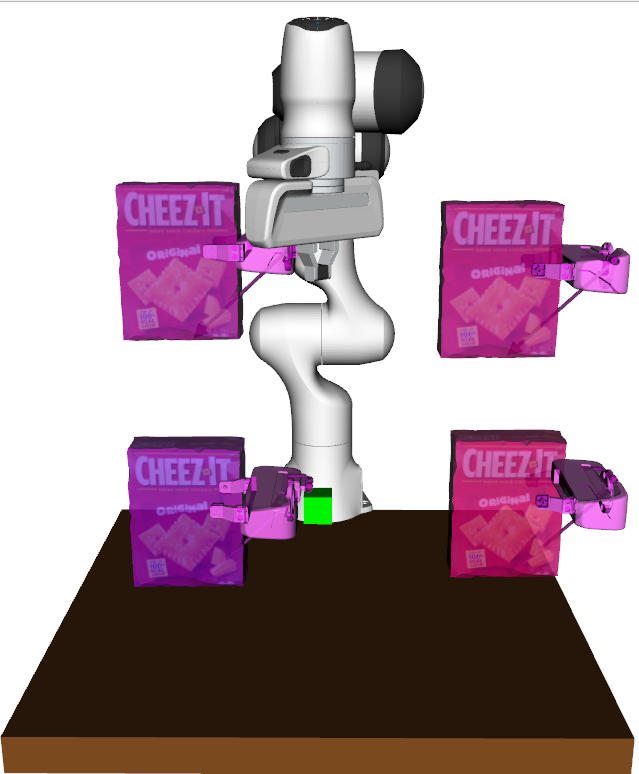} };
\node(1b)[xshift=0cm,yshift=-5cm]{\includegraphics[width=0.23\textwidth]{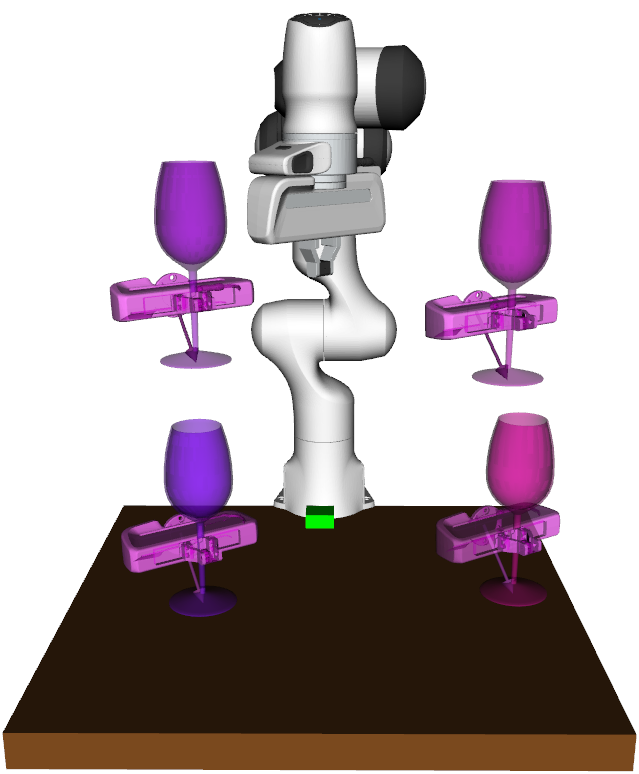} };
\node(1c)[xshift=4cm,yshift=-5cm]{\includegraphics[width=0.23\textwidth]{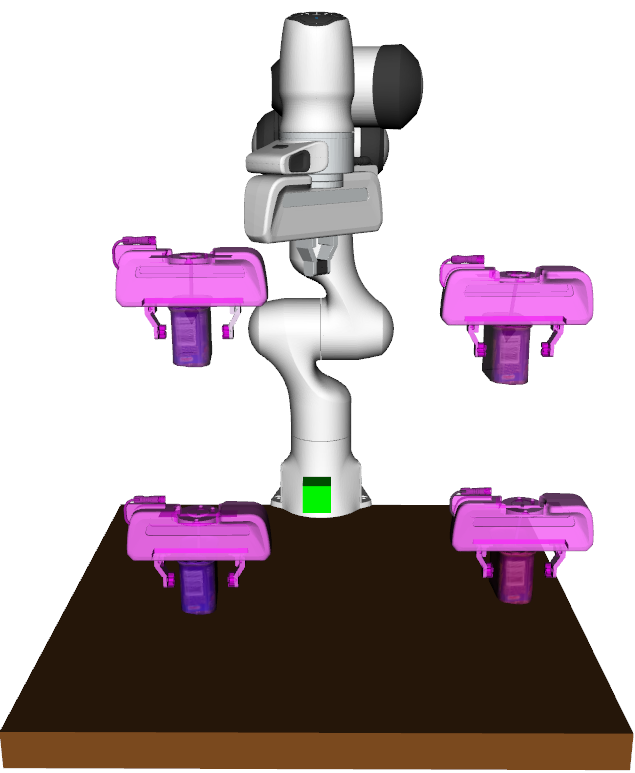} };
\node[fill=white, opacity=0.75,xshift=2.0cm,yshift=-2.25cm]{\color{blue} \small (a)};
\node[fill=white, opacity=0.75,xshift=0.0cm,yshift=-7.25cm]{\color{blue} \small (b)};
\node[fill=white, opacity=0.75,xshift=4.0cm,yshift=-7.25cm]{\color{blue} \small (c)};
\draw[blue, ultra thick, ->] (0.5cm,0cm) to[out=-160,in=110] (-1cm, -3cm);
\draw[blue, ultra thick, ->] (1.0cm,-3.25cm) to[out=0,in=180] (3.5cm, -3.25cm);
\node[ opacity=0.75,xshift=-1.5cm,yshift=-0.5cm, text width=1cm]{\color{blue} \small update\_object service};
\node[ opacity=0.75,xshift=2.5cm,yshift=-3.0cm, text width=3cm]{\color{blue} \footnotesize update\_object service};
\end{tikzpicture}
\vspace*{-7pt}
\caption{Interactive updating of the target object for a generic 4-step task. (a) Original Cheez-It object. (b) After updating the object with a goblet mesh. (c) After calling the \texttt{update\_object} with a Spam CAD model. For (b) and (c) we also sent a grasp query and updated the end effector waypoint}
\label{fig:ui_grasp_tab_3}
\end{figure}

The Affordance Template’s object-updating feature can easily be exploited to reuse a single AT JSON file for multiple objects. Figure \ref{fig:ui_grasp_tab_3}(a) shows a Cheez-It Affordance Template, with the end effector waypoints set appropriately by the grasp planner with a side-grasp. Now imagine that we want to perform this same task (moving an object through the same four poses in space) but instead of using the Cheez-It box, we want to use different objects.

We can make use of the update object function  to try out diverse objects.
Figures \ref{fig:ui_grasp_tab_3}(b) and (c) show the visualization of the same AT trajectory, after calling the service and updating
the object with a wine glass CAD model and a Spam tin mesh. An important observation to make here
is the fact that the service call updates only the display object, not directly the Object Waypoints. The Object Waypoints are nonetheless updated because the whole Affordance Template is rebuilt after the    update object function is called, hence the Object Waypoints are updated down the line.
Furthermore, observe that the grasps shown in Figure \ref{fig:ui_grasp_tab_3} for the new objects are not the same grasps as the ones used for the Cheez-It box. The original EE waypoints were no longer viable to use for the updated objects, as their shape is obviously different. The waypoints shown were obtained by making a call to the grasp planner (using the \texttt{Get grasps} button in the Grasping Tab) and selecting one of the grasps returned by the planner\footnote{Video showing the update object feature is available here: \url{https://youtu.be/C1axyAKVkZE}}.


\subsection{ROI Tab}
In many applications, the geometry of the object to be manipulated might
not be known beforehand. Sensor data, such as point clouds, is used to
describe the object's geometry and as input to the grasp planning pipeline.
Such point cloud must be previously segmented such that only the points corresponding
to the object are considered.

A user might have available tools to perform automatic object segmentation under certain circumstances (e.g., tabletop segmentation), however this is not always the case. To address these scenarios, we developed the \texttt{ROI Tab}, which is a widget that offers the user an easy way to interactively select a region of the point cloud to be segmented and converted to a mesh form, which is then
automatically used to update the object geometry of the currently used Affordance Template.

\begin{figure*}[t!]
\centering
\begin{tikzpicture}[framed,background rectangle/.style={thick,draw=black, rounded corners=1em, inner sep=0.05cm}]
\node(1a)[xshift=0cm,yshift=0cm]{\includegraphics[height=4cm]{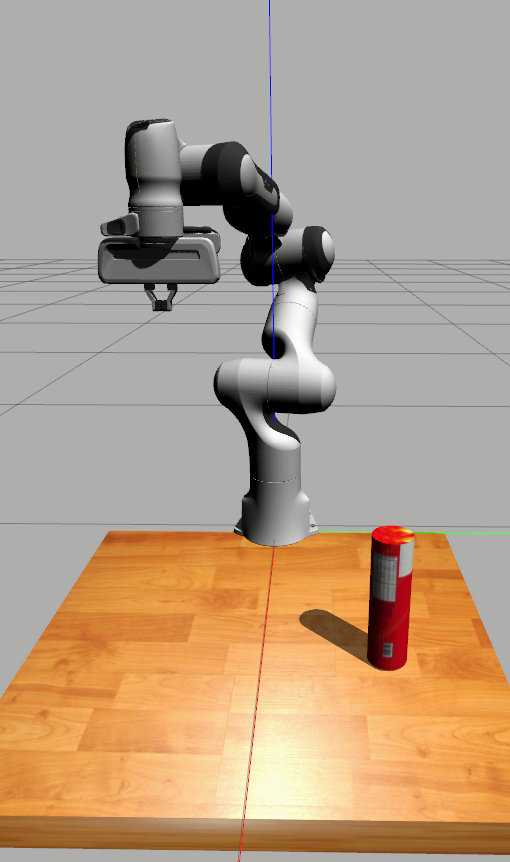} };
\node(1b)[xshift=2.45cm,yshift=0cm]{\includegraphics[height=4cm]{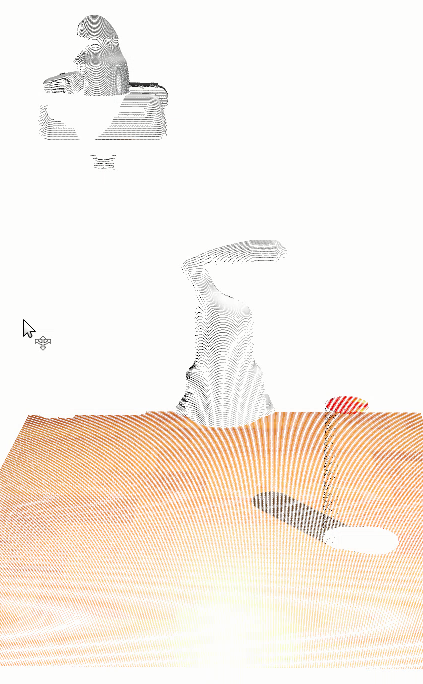} };
\node(1c)[xshift=6.0cm,yshift=0cm]{\includegraphics[height=4cm]{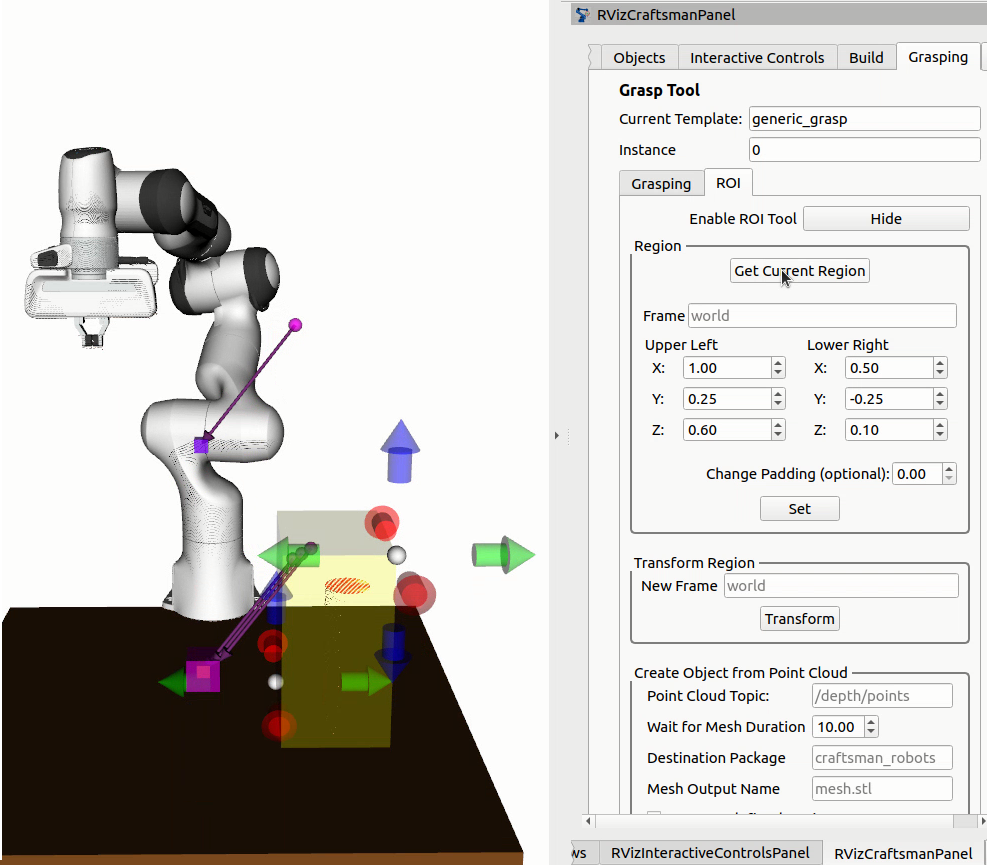} };
\node(1d)[xshift=9.6cm,yshift=0cm]{\includegraphics[height=4cm]{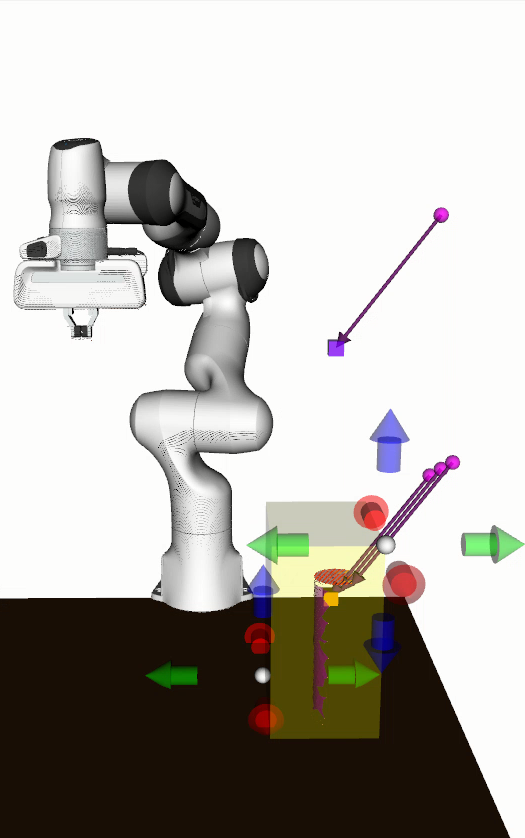} };
\node(1e)[xshift=12.2cm,yshift=0cm]{\includegraphics[height=4cm]{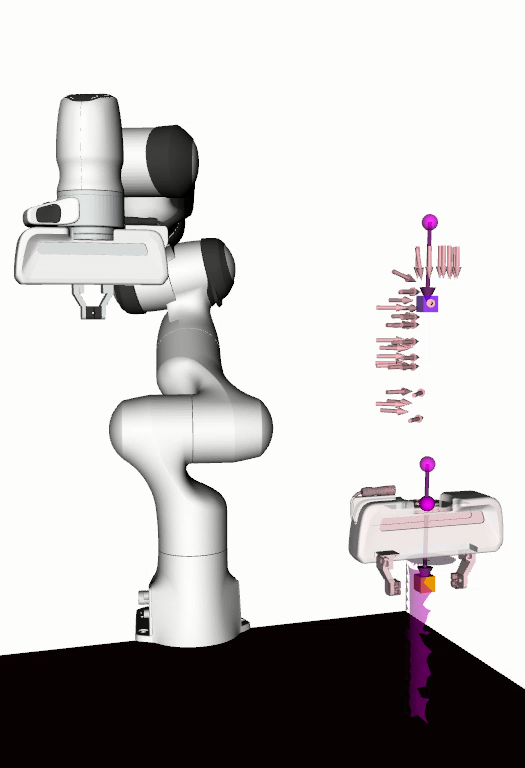} };
\node(1f)[xshift=14.7cm,yshift=0cm]{\includegraphics[height=4cm]{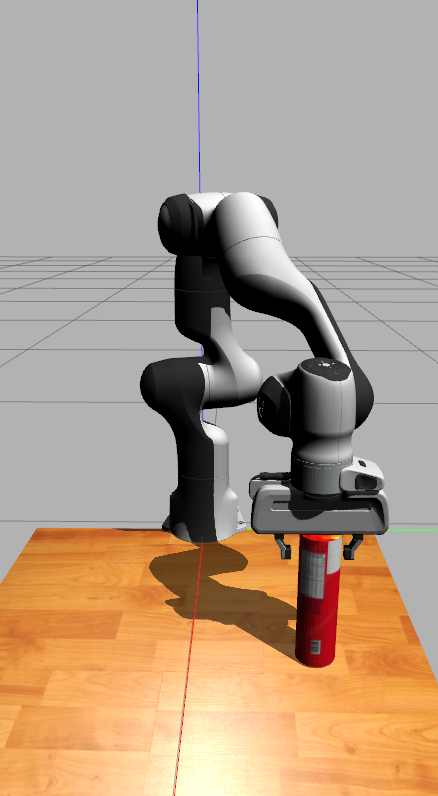} };
\draw [fill=blue,opacity=0.2] (6.4cm,1.40cm) rectangle (8.3cm,-2.0cm);

\node[opacity=0.75,xshift=0.0cm,yshift=1.75cm]{\color{blue} \small (a)};
\node[fill=white, opacity=0.75,xshift=2.5cm,yshift=1.75cm]{\color{blue} \small (b)};
\node[fill=white, opacity=0.75,xshift=5.25cm,yshift=1.75cm]{\color{blue} \small (c)};
\node[fill=white, opacity=0.75,xshift=9.25cm,yshift=1.75cm]{\color{blue} \small (d)};
\node[opacity=0.75,xshift=12.0cm,yshift=1.75cm]{\color{blue} \small (e)};
\node[opacity=0.75,xshift=14.75cm,yshift=1.75cm]{\color{blue} \small (f)};
\end{tikzpicture}
\vspace*{-7pt}
\caption{Pipeline of ROI Tab usage: (a) Original scene, (b) Pointcloud acquired from an overhead sensor, (c) Using the ROI Tab and interactive tools to obtain a segmented volume of the cloud, (d) Using the segmented cloud (converted now into a mesh) to update the current Affordance Template's object, (e) Generating grasps with the Grasping Tab; and (f) Executing the first motion towards the selected grasp.}
\label{fig:ui_roi_tab}
\end{figure*}
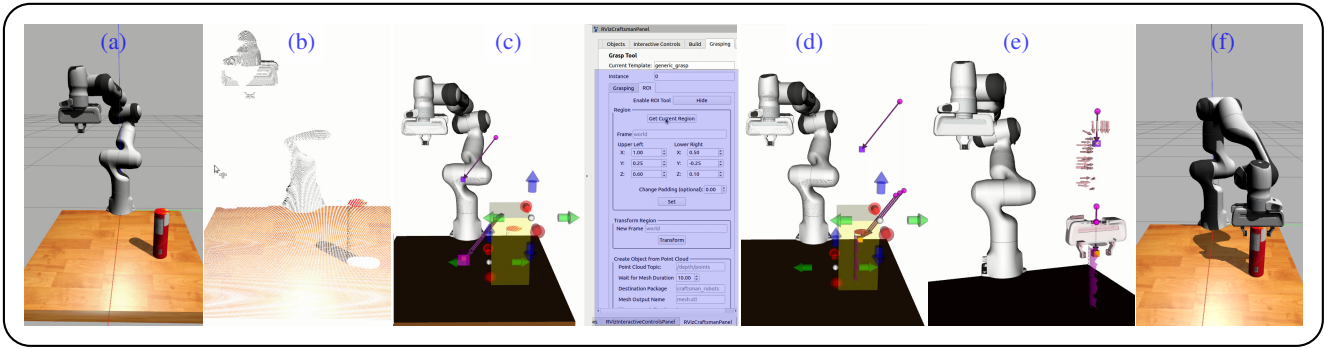

Figure \ref{fig:ui_roi_tab}(b) shows the \texttt{ROI Tab} being used to grasp a Pringles container
using a Panda arm\footnote{Video of the ROI Tab used with the Panda arm is  available here: \url{https://youtu.be/5d5w1vB6wzM}}. Fig. (a) shows the scene as seen in Gazebo and (b) shows the point cloud
obtained from a RGB-D sensor located overhead. Fig. (c) shows the ROI tab (highlighted in blue). It
contains a set of edit boxes that allow a user to define a bounding box to extract a desired volume of the point cloud. The box's dimensions can also be edited using interactive markers in the 3D Rviz window (also shown in Fig(c)). After the user presses the \texttt{Create mesh} button in the ROI tab, two things happen: (1) the point cloud is filtered, and a mesh is created with the segmented points; and (2) the newly created mesh is used to update the target object in the AT currently loaded (Figure (d)). Finally, the user can utilize the Grasping Tab to generate grasps  using the updated mesh and the task steps, which are the same regardless of the change of object geometry (Figure (e)). The last figure of the group shows the Panda arm grasping the object with one of the candidate grasps. 
\section{Examples}
\label{sec:examples}

In this section, we present some examples of using the grasp planning tools developed
by this project with a number of different robots. A playlist with all our demonstrations is
also publicly available\footnote{Playlist available here: \url{https://www.youtube.com/playlist?list=PLmQko6gr2EvHqGXtsES6WOjAiPudvLZ16}}.

\subsection{Handing over a wine glass}
\begin{figure}[H]
\centering
\begin{tikzpicture}[framed,background rectangle/.style={thick,draw=black, rounded corners=1em, inner sep=0.05cm}]
\node[xshift=0cm,yshift=0cm]{\includegraphics[width=0.24\textwidth]{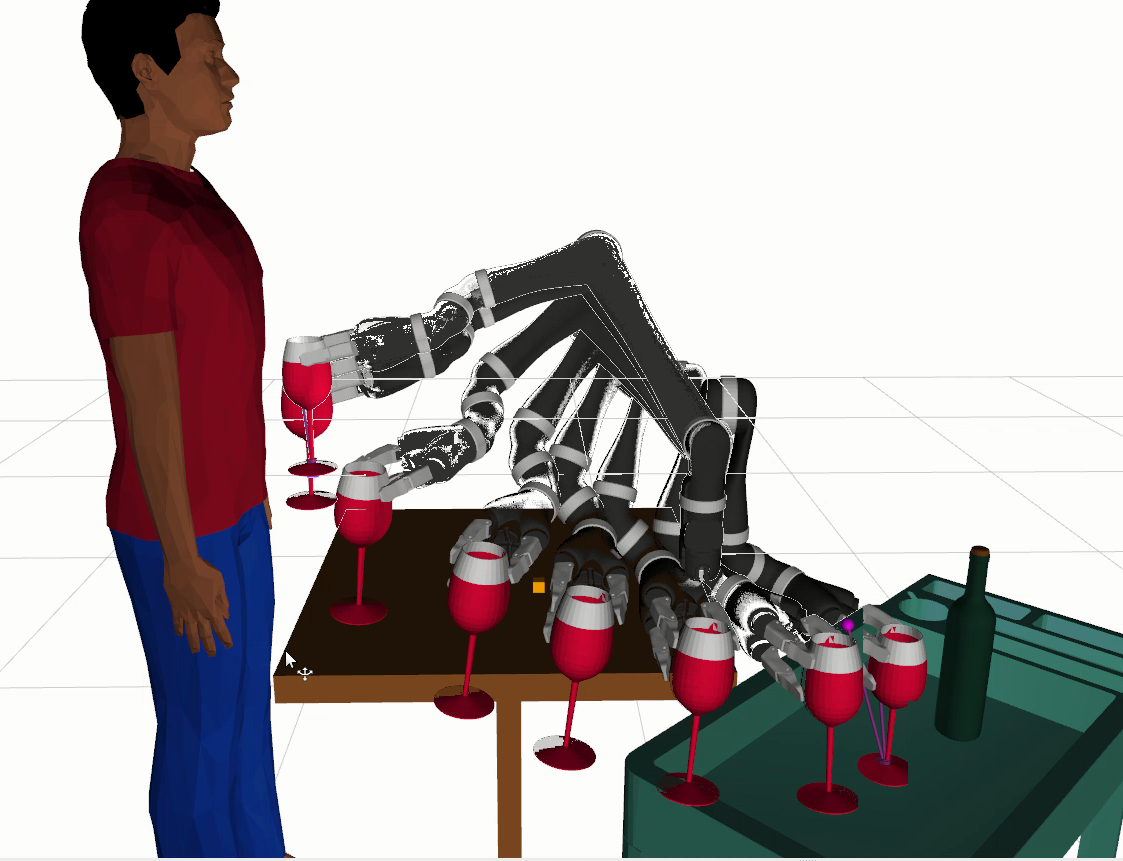} };
\node[xshift=4.0cm,yshift=0cm]{\includegraphics[width=0.22\textwidth]{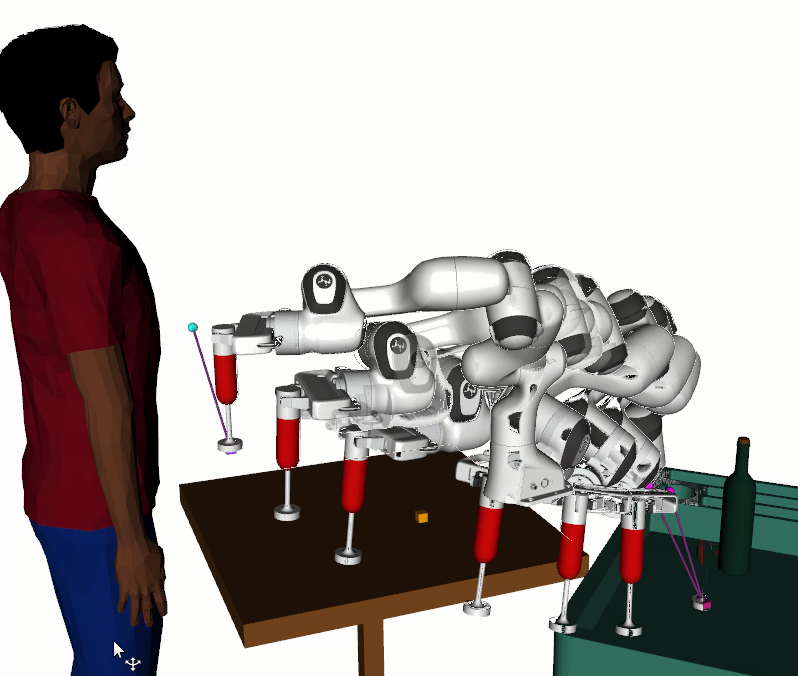} };
\node[fill=white, opacity=0.75,xshift=0cm,yshift=1.4cm]{\color{blue} \scriptsize (a) Jaco};
\node[fill=white, opacity=0.75,xshift=4.0cm,yshift=1.4cm]{\color{blue} \scriptsize (b) Panda};
\end{tikzpicture}
\vspace*{-7pt}
\caption{Jaco (and Panda) manipulators performing a handover task. Overlay of handover motion. Note how the tilting is very small, with the glass being kept upright at most times.}
\label{fig:to7_jaco_wine_2}
\end{figure}

Figure \ref{fig:to7_jaco_wine_2}(a) and (b) shows a Jaco and Panda arm performing a simple handover task consisting of transporting a glass of wine
from a utility cart to be presented to a waiting customer\footnote{Videos of Jaco test: \url{https://youtu.be/bvIVQOjKwS0}. Panda test: \url{https://youtu.be/t4l36rmIY_k}}. In these examples, the task
description used is the same, with the only difference being that the target object was changed
from a wine glass (for Jaco) to a flute (for Panda); this change was necessary as the Panda gripper
opening was not wide enough to hold the original glass. This task had an additional path
constraint of keeping the object with a tilt less than $20^\circ$. 

\subsection{Painting a square on a table}
\begin{figure}[h!]
\centering
\begin{tikzpicture}[framed,background rectangle/.style={thick,draw=black, rounded corners=1em, inner sep=0.05cm}]
\node[xshift=0cm,yshift=0cm]{\includegraphics[width=0.165\textwidth]{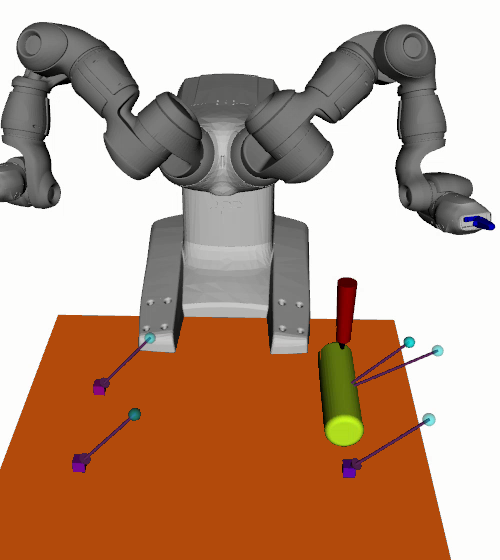} };
\node[xshift=2.775cm,yshift=0cm]{\includegraphics[width=0.165\textwidth]{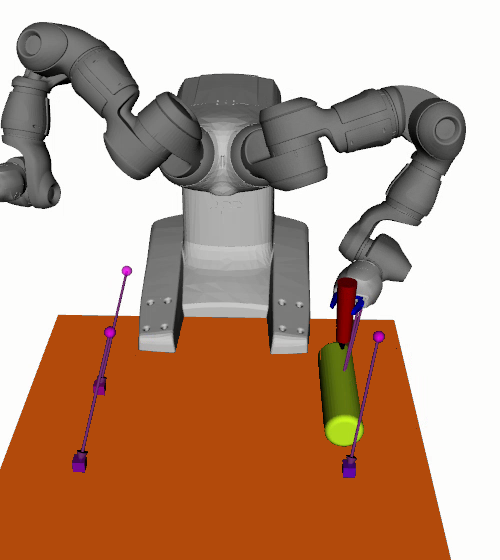} };
\node[xshift=5.5cm,yshift=0cm]{\includegraphics[width=0.165\textwidth]{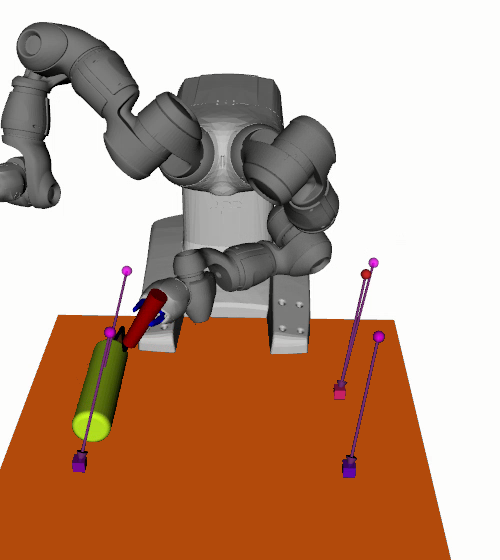} };

\node[fill=white, opacity=0.75,xshift=0cm,yshift=1.5cm]{\color{blue} \scriptsize (a) Start};
\node[fill=white, opacity=0.75,xshift=2.75cm,yshift=1.5cm]{\color{blue} \scriptsize (b) Waypoint 1};
\node[fill=white, opacity=0.75,xshift=5.5cm,yshift=1.5cm]{\color{blue} \scriptsize (c)  Waypoint 2};

\draw[draw, fill=white, opacity=0.25] (0.5cm, -0.5cm) circle (0.5cm);
\node[xshift=0cm,yshift=-3.5cm]{\includegraphics[width=0.165\textwidth]{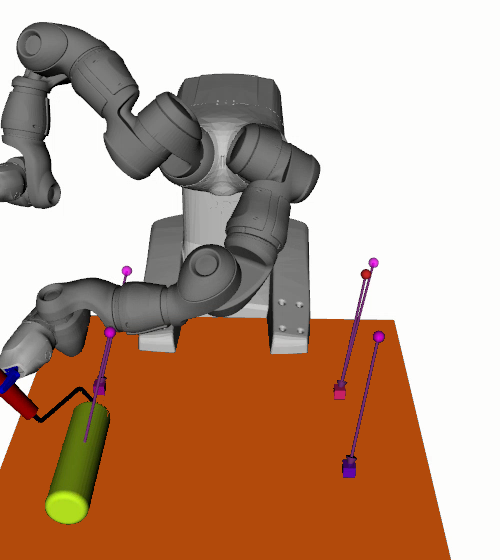} };
\node[xshift=2.75cm,yshift=-3.5cm]{\includegraphics[width=0.165\textwidth]{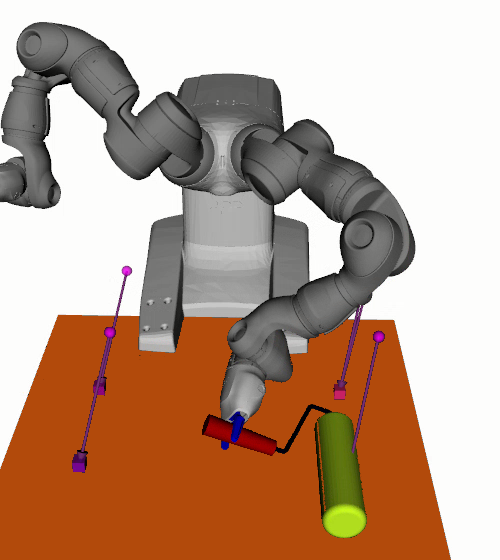} };
\node[xshift=5.5cm,yshift=-3.5cm]{\includegraphics[width=0.165\textwidth]{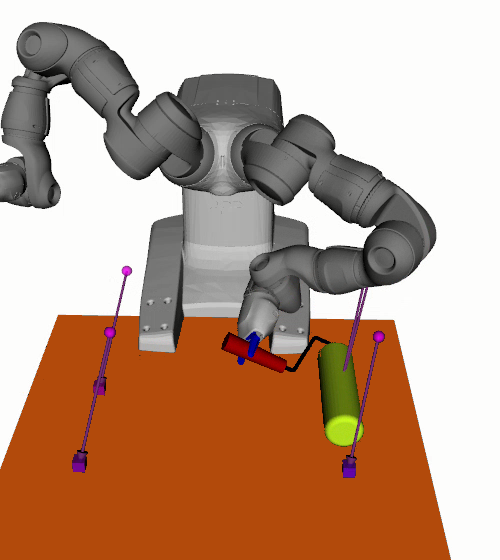} };

\node[            opacity=0.75,xshift=0cm,yshift=-2.0cm]{\color{blue} \scriptsize (d) Waypoint 3};
\node[            opacity=0.75,xshift=2.75cm,yshift=-2.0cm]{\color{blue} \scriptsize (e) Waypoint 4};
\node[fill=white, opacity=0.75,xshift=5.5cm,yshift=-2.0cm]{\color{blue} \scriptsize (f) Waypoint 5};

\end{tikzpicture}
\vspace*{-7pt}
\caption{Yumi robot performing a painting task with 5 object waypoints. The path constraints require the roller to be in contact with the surface at all times, following closely a line, while the handle can rotate with respect to the roller's axis.}
\label{fig:to7_yumi_painting_2}
\end{figure}
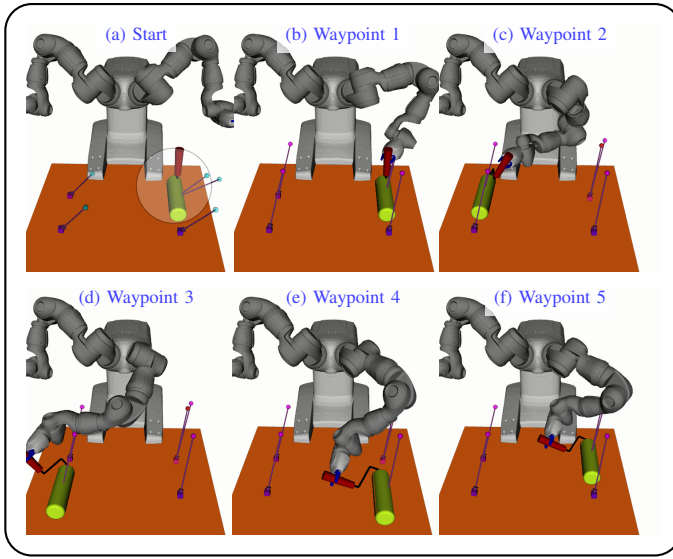

This demonstration showcases the utility of planning with path
constraints and the use of a grasp evaluator metric that can be used for tasks with multiple waypoints. In this
example, the task consists of painting a square on a table (Figure \ref{fig:to7_yumi_painting_2}). The roller has to keep its surface in contact with
the table (Z tolerance is zero). The roller has to closely follow a line between waypoints, so the
tolerances in X and Y are very small during motion (1mm). Finally, the handle can rotate with respect to the roller axis, as long as contact is kept\footnote{Video of Yumi painting test: \url{https://youtu.be/YeGa8pRv6wU}}.

\subsection{Valkyrie walking towards a cart to pick up a beer}
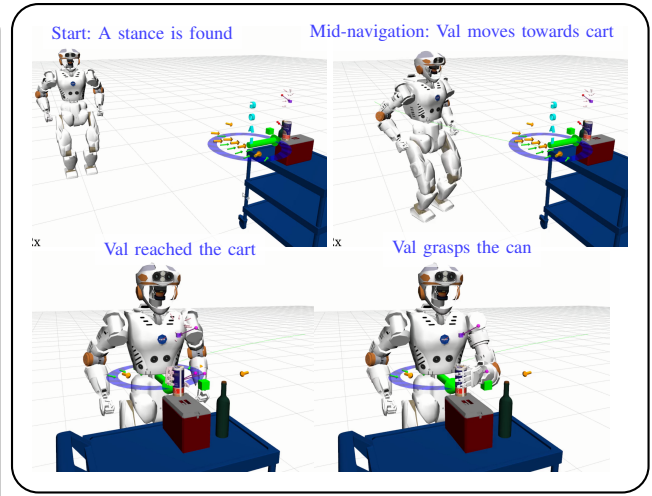
\begin{figure}[h]
\centering
\begin{tikzpicture}[framed,background rectangle/.style={thick,draw=black, rounded corners=1em, inner sep=0.05cm}]
\node[xshift=0cm,yshift=0cm]{\includegraphics[width=0.22\textwidth]{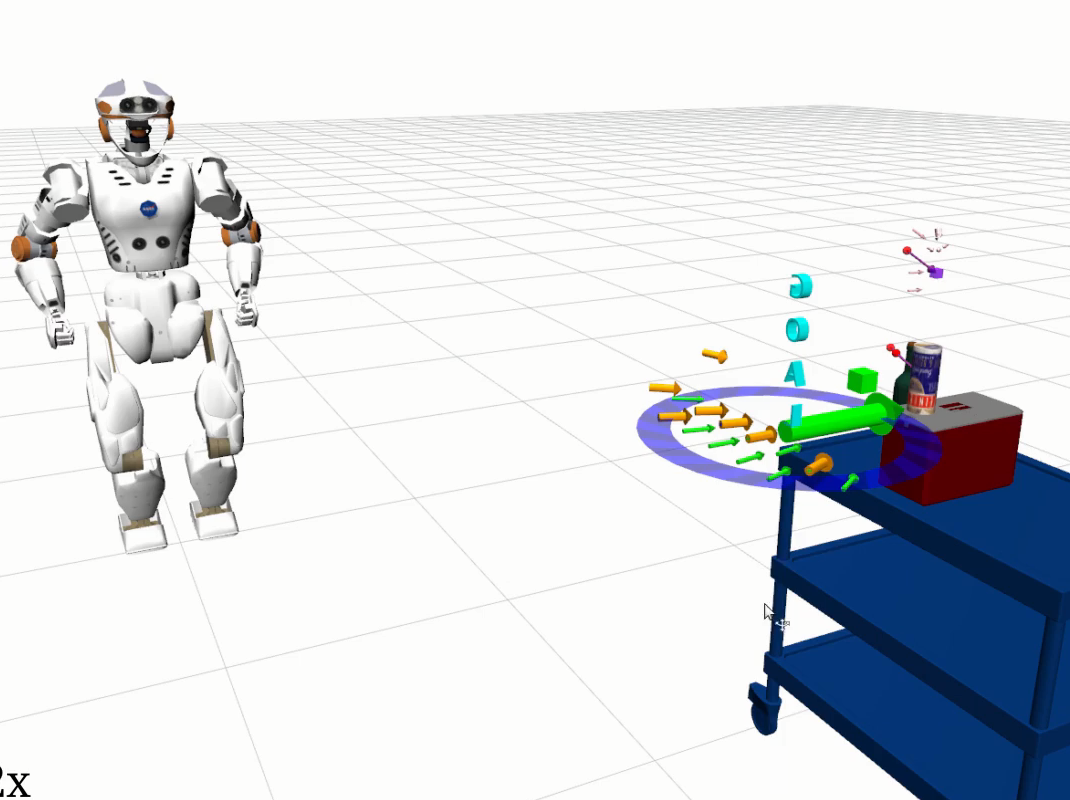} };
\node[xshift=4.0cm,yshift=0cm]{\includegraphics[width=0.22\textwidth]{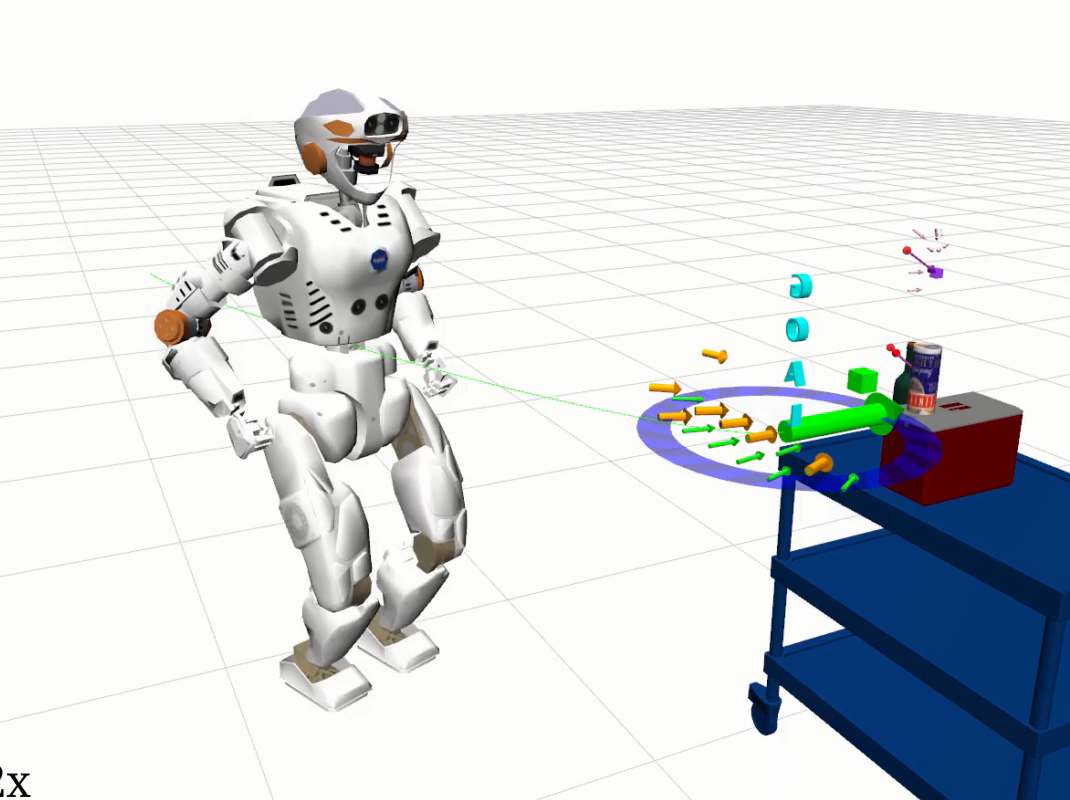} };
\node[fill=white, opacity=0.75,xshift=-0.5cm,yshift=1.375cm]{\color{blue} \scriptsize Start: A stance is found};
\node[fill=white, opacity=0.75,xshift=3.75cm,yshift=1.375cm]{\color{blue} \scriptsize Mid-navigation: Val moves towards cart};

\node[xshift=0cm,yshift=-3.0cm]{\includegraphics[width=0.22\textwidth]{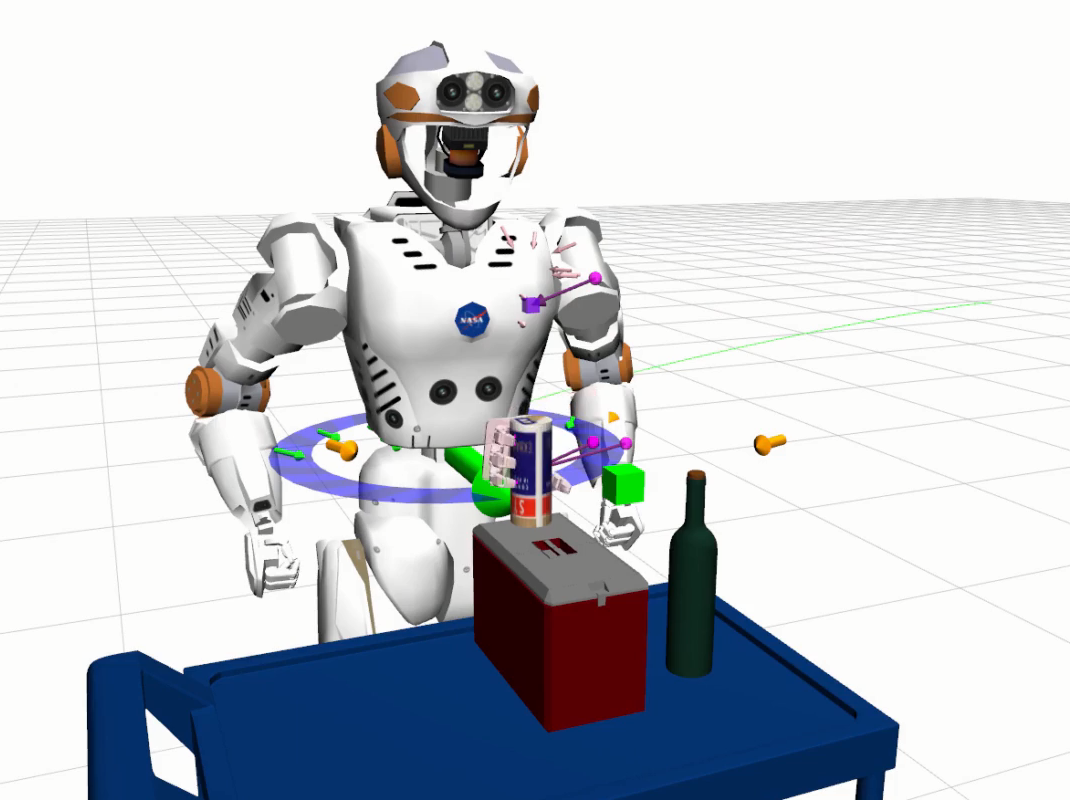} };
\node[xshift=3.75cm,yshift=-3.0cm]{\includegraphics[width=0.22\textwidth]{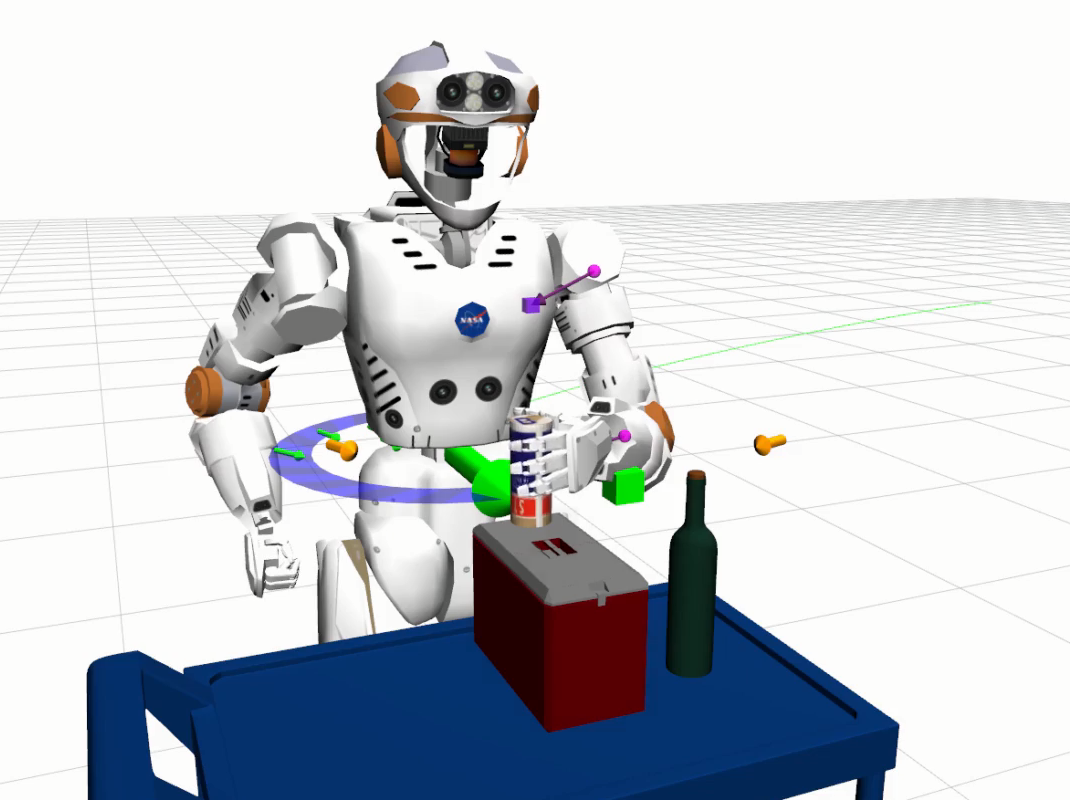} };
\node[fill=white, opacity=0.75,xshift=0cm,yshift=-1.5cm]{\color{blue} \scriptsize Val reached the cart};
\node[fill=white, opacity=0.75,xshift=3.75cm,yshift=-1.5cm]{\color{blue} \scriptsize Val grasps the can};

\end{tikzpicture}
\vspace*{-5pt}
\caption{Example of grasp planning integration with stance generation in CRAFTSMAN. Snapshots show Valkyrie planning simultaneously a robot placement that allows it to reach a number of candidate grasps to grab the beverage can.}
\label{fig:to7_val_coors_1}
\end{figure}

In this demonstration (Figure \ref{fig:to7_val_coors_1}), we use the mobile grasping capabilities developed (combining the grasp planning features with the existing stance generation module)  and use them
to enable Valkyrie to plan both a robot stance and a set of grasp solutions to pick up a drink can
located outside of its reachable space\footnote{Video of Valkyrie pick: \url{https://youtu.be/PraEhCeglRo}}. 

\subsection{Fetch navigating towards a cart to pour a drink}
\begin{figure*}[t]
\centering
\begin{tikzpicture}[framed,background rectangle/.style={thick,draw=black, rounded corners=1em, inner sep=0.05cm}]
\node[xshift=0cm,yshift=0cm]{\includegraphics[height=4cm]{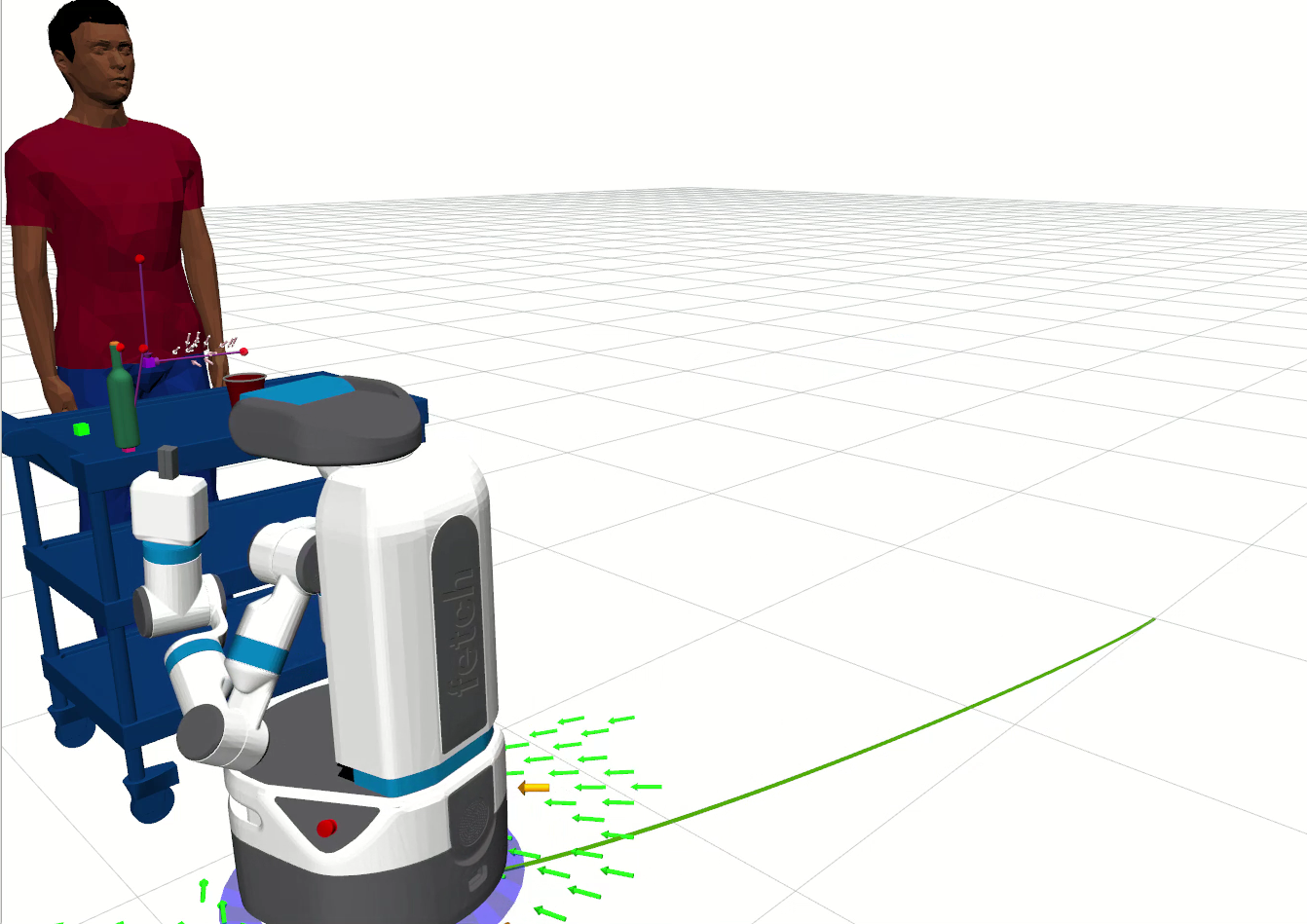} };
\node[xshift=4.85cm,yshift=0cm]{\includegraphics[height=4cm]{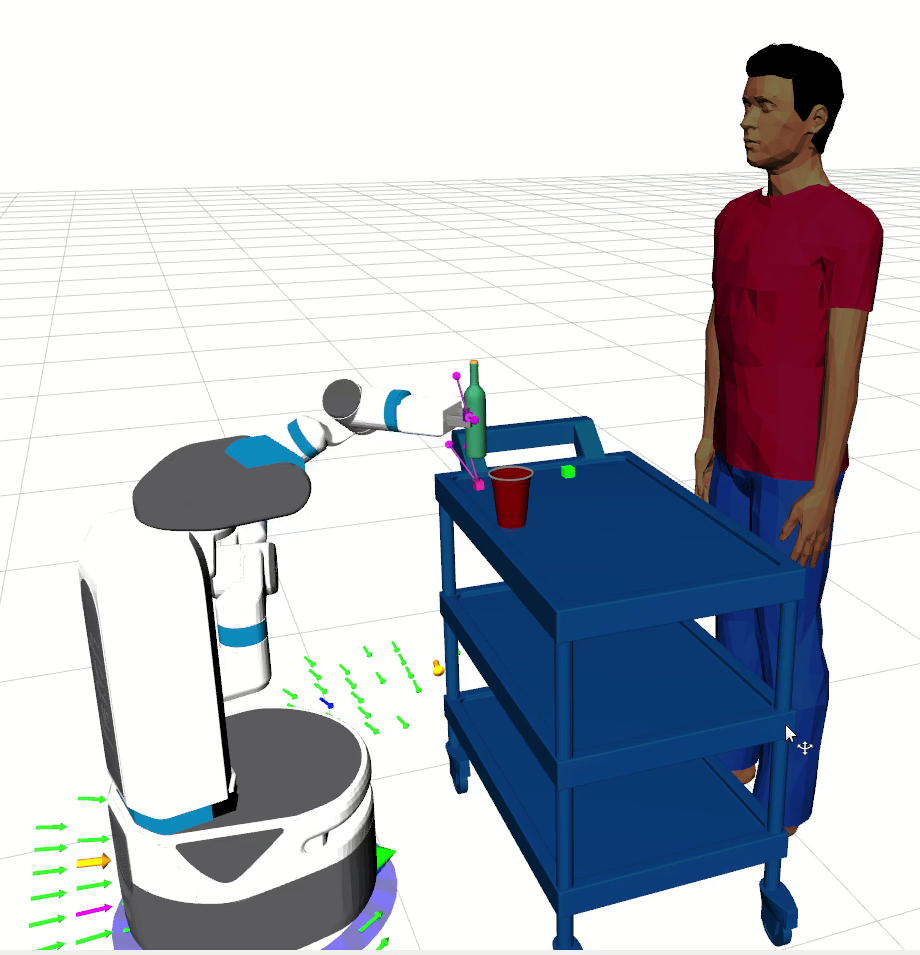} };
\node[xshift=8.8cm,yshift=0cm]{\includegraphics[height=4cm]{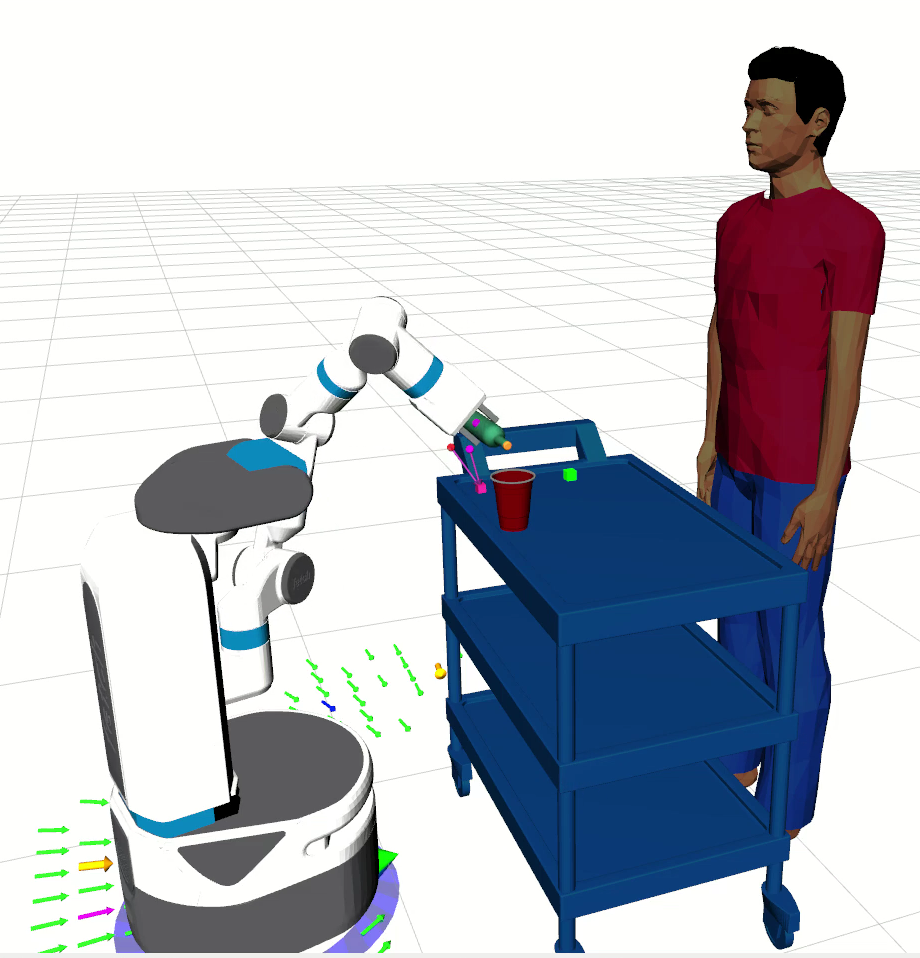} };

\node[fill=white, opacity=0.75,xshift=0.0cm,yshift=2.0cm]{\color{blue} \scriptsize (a) Fetch navigates towards the cart};
\node[fill=white, opacity=0.75,xshift=4.5cm,yshift=2.0cm]{\color{blue} \scriptsize (b) Fetch picks the bottle};
\node[fill=white, opacity=0.75,xshift=8.8cm,yshift=2.0cm]{\color{blue} \scriptsize (c) Fetch completes the pouring task};

\end{tikzpicture}
\vspace*{-5pt}
\caption{Fetch robot performing a (mobile) pouring task on a cart located 3 meters away.}
\label{fig:to7_fetch_pour_left}
\end{figure*}
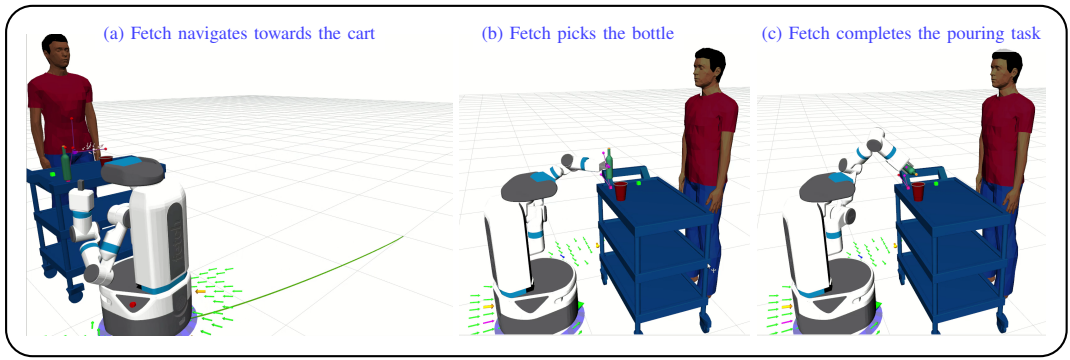

This demonstration showcases the Fetch robot performing a drink pouring task. The additional constraint here is that
the service cart where the drink is located is far from its reachable space. Figure \ref{fig:to7_fetch_pour_left} shows
some snapshots of the task \footnote{Video of the Fetch pouring test: \url{https://youtu.be/gZKIBqDPFUE}}.
The pouring task is defined with 3 object waypoints, the first for reaching the object, the second for raising the bottle to
be close to the cup to receive the liquid, and the final object waypoint is a rotation of the second one by $90^{\circ}$, such that the wine bottle is tilted over the cup. The end effector waypoints are 5, with 3 matching the 3 object waypoints and 2 additional pre-grasp waypoints at the start of the task.

\section{Conclusions and Future Work}
\label{sec:future_work}
In this paper we have presented ADAMANT, the grasp software module developed to provide the CRAFTSMAN framework with grasp
planning capabilities. Our implemented modules allow a user to generate a set of candidate grasps for a manipulation task, ordered in such a way that the best grasps are presented first. The software tools presented were developed
using a plugin-based design, such that they are easily extensible to use existing third-party libraries for grasp
generation. Our tools also include interactive tools that allow the user to update the object being manipulated
on the fly, either with an existing CAD model or with sensor data. We provided multiple examples showcasing the
usefulness of our enhancements.

As future work, we'd like to explore how to better integrate perception tools into our interactive tools (e.g., the \texttt{ROI Tab}). Algorithms for object completion (to address partial point clouds) would be particularly useful.
Additionally, the integration of algorithms to segment an object into parts would be helpful to
 improve the grasp generation process.

\bibliographystyle{IEEEtran}
\normalsize
\bibliography{smart_planning}

\end{document}